\documentclass[10pt,twocolumn,letterpaper]{article}

\usepackage{cvpr}              
\usepackage{algorithm}
\usepackage{algpseudocodex}
\usepackage{amsmath}
\usepackage{amssymb}
\usepackage{multirow}
\usepackage{booktabs}
\usepackage{makecell}
\usepackage{subcaption}
\usepackage{fancybox}
\usepackage{fancyvrb}
\usepackage{bm}
\usepackage{multirow}
\usepackage{booktabs}
\usepackage{bbding}
\usepackage{makecell}
\usepackage{subcaption}
\usepackage{colortbl}
\usepackage{xcolor}
\usepackage{graphicx} 
\usepackage{calligra}
\usepackage{wrapfig}
\usepackage{tabularx}
\definecolor{softred}{RGB}{255, 178, 178}  
\definecolor{softorange}{RGB}{255, 218, 179} 
\definecolor{softyellow}{RGB}{255, 244, 191} 
\newcommand{\dd}[2]{$#1\scriptstyle{\pm#2}$}
\newcommand{\ddbf}[2]{$\mathbf{#1\scriptstyle{\pm#2}}$}

\definecolor{cvprblue}{rgb}{0.21,0.49,0.74}
\usepackage[pagebackref,breaklinks,colorlinks,allcolors=cvprblue]{hyperref}

\def\paperID{2459} 
\def\confName{CVPR}
\def\confYear{2026}

\title{ActiveVLA: Injecting Active Perception into Vision-Language-Action Models for Precise 3D Robotic Manipulation}

\author{Zhenyang Liu$^{1,2}$, Yongchong Gu$^{1}$, Yikai Wang$^{3}$\footnotemark[1], Xiangyang Xue$^{1}$\footnotemark[2], Yanwei Fu$^{1,2}$\footnotemark[2]\\
$^1$\normalsize Fudan University
$^2$\normalsize Shanghai Innovation Institute
$^3$\normalsize Nanyang Technological University\\
{\tt\small liuzy24@m.fudan.edu.cn, yongchonggu22@m.fudan.edu.cn, yikai.wang@ntu.edu.sg,} \\
{\tt\small xyxue@fudan.edu.cn, yanweifu@fudan.edu.cn}\\
Project Page: \href{https://zhenyangliu.github.io/ActiveVLA/}{ZhenyangLiu.github.io/ActiveVLA}}

\begin{document}
\maketitle
\begin{abstract}
Recent advances in robot manipulation have leveraged pre-trained vision-language models (VLMs) and explored integrating 3D spatial signals into these models for effective action prediction, giving rise to the promising vision-language-action (VLA) paradigm. However, most existing approaches overlook the importance of active perception: they typically rely on static, wrist-mounted cameras that provide an end-effector-centric viewpoint. As a result, these models are unable to adaptively select optimal viewpoints or resolutions during task execution, which significantly limits their performance in long-horizon tasks and fine-grained manipulation scenarios. To address these limitations, we propose \textbf{ActiveVLA}, a novel vision-language-action framework that empowers robots with active perception capabilities for high-precision, fine-grained manipulation. ActiveVLA adopts a coarse-to-fine paradigm, dividing the process into two stages:
(1) Critical region localization. ActiveVLA projects 3D inputs onto multi-view 2D projections, identifies critical 3D regions, and supports dynamic spatial awareness.
(2) Active perception optimization. Drawing on the localized critical regions, ActiveVLA uses an active view selection strategy to choose optimal viewpoints. These viewpoints aim to maximize amodal relevance and diversity while minimizing occlusions. Additionally, ActiveVLA applies a 3D zoom-in to improve resolution in key areas. Together, these steps enable finer-grained active perception for precise manipulation.
Extensive experiments demonstrate that ActiveVLA achieves precise 3D manipulation and outperforms state-of-the-art baselines on three simulation benchmarks. Moreover, ActiveVLA transfers seamlessly to real-world scenarios, enabling robots to learn high-precision tasks in complex environments.
\end{abstract}

\section{Introduction}

\begin{flushleft}
\textit{“Perception is not a passive process. It’s an active process of hypothesis testing.”}
\hfill -- Richard Gregory
\end{flushleft}

\noindent Pretrained vision-language models (VLMs)~\cite{beyer2024paligemma, wang2024qwen2, bai2025qwen2, karamcheti2024prismatic} have become a highly effective approach for building large-scale vision-language-action (VLA) architectures. Such models have demonstrated impressive generalization and robustness in robotic manipulation~\cite{kimopenvla, black2410pi0, liu2025trivla, intelligence2025pi_, li2023vision, brohan2023rt}. However, most current VLA approaches primarily process 2D visual inputs, requiring massive datasets to bridge the gap between perception and action. In contrast, 3D-aware robotic policies leverage structural cues from 3D data, achieving better sample efficiency and spatial reasoning for complex manipulation tasks~\cite{shridhar2023perceiver, 3d-da, gervet2023act3d, goyal2023rvt, goyal2024rvt}.

Gregory’s insight captures a core challenge in embodied intelligence: perception must not remain a passive receiver of sensory input but instead become an active hypothesis-testing process—seeking, selecting, and verifying information relevant to the task at hand. However, existing VLA models~\cite{kimopenvla, liu2025trivla, zhen20243d, qu2025spatialvla} largely rely on static or wrist-mounted cameras, constraining observations to a fixed, end-effector-centric viewpoint. Such a setup inherently limits perceptual flexibility: the agent cannot adaptively and dynamically adjust its viewpoint or camera resolution according to the task context. As a result, the absence of adaptive viewpoint selection during long-horizon or fine-grained manipulation prevents current systems from acquiring essential contextual information, thereby undermining the robustness and generalizability of learned policies~\cite{liu2025trivla, shridhar2023perceiver, 3d-da, gervet2023act3d, goyal2023rvt, goyal2024rvt}. Addressing this limitation is crucial for developing embodied agents capable of adaptive and reliable interaction in complex, real-world environments.

To address this limitation, we propose \textbf{ActiveVLA}, a novel vision-language-action framework that explicitly integrates active perception into robotic manipulation. ActiveVLA enables robots to adaptively and dynamically adjust their viewpoint and camera resolution according to the task context, allowing more informative observations to be acquired on demand. The framework equips robots with two complementary capabilities:
(1) \textbf{Active viewpoint selection}, which autonomously determines optimal camera perspectives during task execution to maximize visibility and task relevance while minimizing occlusions; and
(2) \textbf{Active 3D zoom-in}, which identifies and selectively enhances high-resolution views of task-critical regions within the 3D scene.
By dynamically refining its perceptual input through these mechanisms, ActiveVLA enables more precise, context-aware, and reliable action prediction, ultimately achieving superior adaptability and performance in complex, long-horizon manipulation scenarios.

As illustrated in Figure~\ref{teaser}, ActiveVLA adopts a coarse-to-fine active perception design that integrates 3D spatial reasoning with vision-language understanding. In the \textit{coarse stage}, ActiveVLA projects the 3D point-cloud observations into multiple 2D orthographic views~\cite{goyal2023rvt, goyal2024rvt}, aligning them with the pretrained VLM backbone. By leveraging structural priors from the 3D inputs, the model effectively localizes salient and task-relevant regions in 3D space. In the \textit{fine stage}, ActiveVLA performs active view selection to determine optimal camera poses centered on these key regions, maximizing amodal relevance and spatial diversity while reducing occlusions. It then applies an active 3D zoom-in strategy to enhance the spatial resolution of these areas, enabling precise manipulation of fine-grained details. This closed-loop, coarse-to-fine perception-action pipeline allows ActiveVLA to dynamically adapt its sensory inputs and maintain high effectiveness across complex, multi-step, and long-horizon manipulation tasks.

Comprehensive experiments validate the advantages of ActiveVLA. On RLBench~\cite{james2020rlbench}, it achieves an average success rate of 91.8\%, with some tasks reaching a perfect success rate of 100\%. On COLOSSEUM~\cite{pumacay2024colosseum}, it achieves the highest success rate of 78.3\% in challenging generalization scenarios. ActiveVLA consistently outperforms all baselines on GemBench~\cite{garcia2024towards}, demonstrating superior adaptability across diverse tasks. Real-world evaluations further confirm its robustness and strong generalization to previously unseen configurations. These results demonstrate that integrating active perception into vision-language-action learning is a key step toward adaptive embodied intelligence.

The contributions of this paper are summarized:
\begin{itemize}
\item \textit{\textbf{Active Perception for Vision-Language-Action Models}}: We propose ActiveVLA, a novel vision-language-action framework that equips robots with active perception capabilities, enabling adaptive viewpoint selection and zoom-in mechanisms for precise, fine-grained manipulation.
\item \textit{\textbf{A Novel ActiveVLA Framework}}: ActiveVLA designs a novel coarse-to-fine pipeline that projects 3D point clouds into multi-view 2D representations, predicts key regions via heatmaps, and selectively refines observations for precise, fine-grained manipulation.
\item \textit{\textbf{State-of-the-Art Performance and Real-World Generalization}}: Extensive experiments on RLBench, COLOSSEUM, and GemBench demonstrate that ActiveVLA outperforms state-of-the-art baselines. Real-world robot evaluations show strong generalization and high success rates, highlighting the practical impact of active perception in long-horizon and precision-critical tasks.
\end{itemize}

\section{Related Work}
\noindent\textbf{Vision-Language-Action Models.}
Recent research has increasingly focused on developing generalist robot policies trained on large-scale robotic learning datasets~\cite{o2023open-x, khazatsky2024droid, fang2023rh20t, dasari2024ditpolicy, lin2024datascalinglawsimitation}. Vision-Language-Action (VLA) models have emerged as a promising approach for training such policies~\cite{kim24openvla, embodiedcot, [pi0, wen2025dexvla, pertsch2025fast, niu2024llarva, diffusion-policy, zhou2025chatvla, ding2025humanoid, zhao2025vlas, ding2024quar, tong2024quart, liu2025spatial, wen2024tinyvla}. VLAs extend vision-language models (VLMs), which are pre-trained on massive internet-scale image and text datasets~\cite{zhu2024mipha, zhao2024cobra, zhu2024llavaphi, karamcheti2024prismatic, wang2024qwen2, lu2024deepseek-vl, llava, llava1.5, abdin2024phi3, chen2024internvl}, to the domain of robotic control~\cite{wang2024distrl}. This approach provides several key advantages: leveraging large-scale vision-language model backbones with billions of parameters enables effective learning from vast robotic datasets, while reusing pre-trained weights from internet-scale data enhances VLAs’ ability to interpret diverse language commands and generalize to novel objects and environments, making them adaptable for real-world robotic applications.
ActiveVLA is a vision-language-action (VLA) framework that employs large-scale vision-language model backbones. Unlike previous methods, ActiveVLA further leverages the spatial structure of 3D inputs and utilizes active perception through novel view synthesis of 3D representations, resulting in greater adaptability and improved performance in manipulation tasks.

\begin{figure*}
	\centering
    \includegraphics[width=0.98\linewidth]{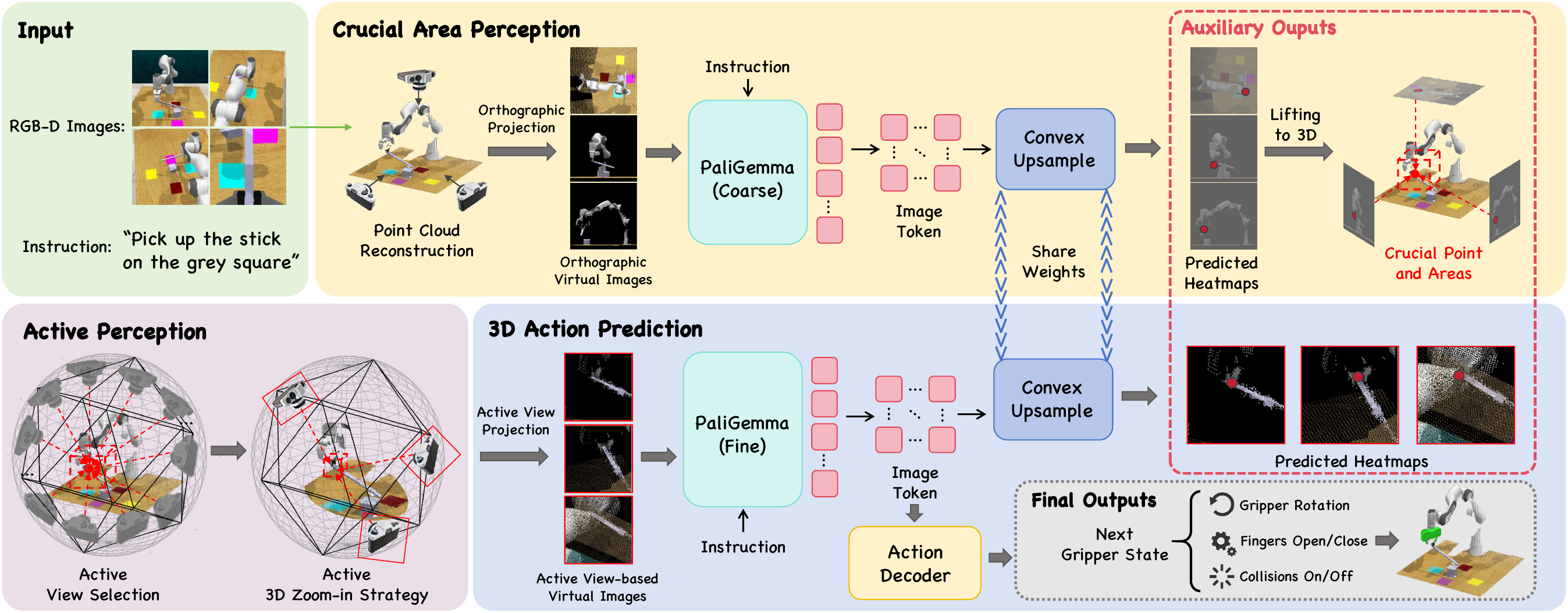}
	\caption{\textbf{The pipeline of ActiveVLA.} \textbf{ActiveVLA} is a 3D vision-language-action framework that adopts a two-stage, coarse-to-fine strategy. In the \emph{coarse stage}, three orthographic projections of the 3D scene and a language instruction are processed by the PaliGemma backbone to generate 2D heatmaps, which are then back-projected to locate the most relevant 3D region. In the \emph{fine stage}, an active perception module selects new views and performs a 3D zoom-in on this region. The refined PaliGemma then predicts heatmaps for key end-effector positions, while an action decoder outputs the final 3D action.
	\label{fig:overall} }
    \vspace{-10pt}
\end{figure*}

\noindent\textbf{3D Modalities in Robotic Learning.} Although VLA models with 2D inputs have been extensively explored, incorporating 3D modalities into robot learning is still an emerging area of research~\cite{zhen20243d, jia2024lift3d, liu2025reasongrounder, yang2025fp3, liu2025neural, li2025pointvla}. Recent studies have started leveraging richer spatial information to enhance robot perception and manipulation capabilities. Some approaches, such as 3DVLA~\cite{leo3d}, present holistic models for unified 3D tasks like generalization, visual question answering, 3D perception, and control. Other methods aim to improve the performance of foundational 2D models by integrating three-dimensional features. Lift3D~\cite{jia2024lift3d} augments 2D vision backbones like DINOv2~\cite{oquab2023dinov2} with both implicit and explicit 3D representations. PointVLA~\cite{li2025pointvla} utilizes separate encoders for 2D images and 3D point clouds, merging their features in a downstream action predictor to inform decision-making. SpatialVLA~\cite{qu2025spatialvla} introduces a specialized position encoding (Ego3D) to inject spatial information into 2D observations.
However, most existing methods lack perceptual flexibility because the models cannot dynamically adjust their viewpoint or camera resolution. As a result, they fail to leverage the advantages of active, egocentric perception in robotic manipulation.
In contrast, ActiveVLA uses an active camera selection strategy to choose optimal viewpoints and an active 3D zoom-in mechanism to enhance point cloud resolution in key regions, leading to better performance in complex, long-horizon tasks.

\section{Our Proposed ActiveVLA}
\noindent \textbf{VLA Setup.} 
Vision-Language-Action models aim to learn a generalizable and robust multi-task robot manipulation policy $\pi$. $\pi$ that maps an observation $\mathbf{o}$ and a language instruction $l$ to an action $\mathbf{a}$:
\begin{equation}
    \pi: (\mathbf{o}, l) \mapsto \mathbf{a},
\end{equation}
The action $\mathbf{a}$ consists of a 6-DoF end-effector pose $T\in SE(3)$, a gripper state $g\in\{0, 1\}$, and a collision flag $c\in\{0, 1\}$ of the next key frame. This setting assumes access to expert demonstrations $\mathcal{D} = \{\tau^i\}_{i=1}^N$ containing $N$ trajectories, where each trajectory contains a language instruction and a sequence of observation-action pairs, \textit{i.e.}, $\tau^{i} = \{l^{i}, (\mathbf{o}_{1}^{i}, \mathbf{a}_{1}^{i}), ..., (\mathbf{o}_{H}^{i}, \mathbf{a}_{H}^{i})\}$.
The observation $\mathbf{o}$ contains RGB-D images captured from one or multiple fixed, end-effector-centric viewpoints. However, this setup limits perceptual flexibility by preventing dynamic viewpoint or resolution adjustment, hindering context acquisition and policy generalization.

\noindent \textbf{ActiveVLA.} 
As shown in Figure~\ref{fig:overall}, ActiveVLA addresses the limitation of perceptual flexibility through a coarse-to-fine active perception framework that adaptively adjusts the camera’s viewpoint and zoom-in level according to the task context, thereby enhancing scene exploration and facilitating fine-grained interaction.  By integrating 3D spatial reasoning with vision-language understanding, our approach progressively refines perceptual focus from a broad overview to task-critical regions, enhancing both contextual awareness and generalization.
ActiveVLA enables robots to attend to critical 3D regions and actively sense the environment, allowing adaptive viewpoints and zoom-in for fine-grained manipulation.

\subsection{3D Crucial Area Perception}
During each task, ActiveVLA first identifies the core region in the 3D scene to serve as the focus for active perception.

\noindent \textbf{Multi-View Rendering.} Given RGB-D images as visual input, ActiveVLA reconstructs a point cloud of the scene using images captured by calibrated cameras. To match the 2D image input required by the VLM backbone, we render three orthographic projection images of the point cloud from top, front, and right viewpoints. Specifically, for each view, we render three image maps with a total of 7 channels: RGB, depth, and the coordinates of the points in the world frame (3 channels). The coordinates help establish the correspondence of pixels across views, \textit{i.e.}, if pixels from different views share the same $(x, y, z)$, they correspond to the same point in 3D. We use PyTorch3D~\cite{ravi2020pytorch3d} for rendering. The rendered image at viewpoint \(v\) is given by:
\begin{equation}
I^{(v)}(u_x, u_y) = \sum_{i=1}^N \mathbf{c}_i \cdot \delta\big( (u_x, u_y) - \pi^{(v)}(\mathbf{p}_i) \big),
\end{equation}
where \(\pi^{(v)}(\cdot)\) denotes the orthographic projection. The rendered color at each pixel is taken from the point with the minimal depth \(z_i^{(v)}\) among all points projecting to that pixel, ensuring correct occlusion handling.

\noindent \textbf{3D Crucial Area Extraction.}
These images are then used as input for the VLM backbone, which is designed to process images along with accompanying text prompts. Since the VLM’s global representations are not sufficient for precise spatial localization, we introduce a heatmap prediction module to recover fine-grained spatial attention and identify object positions within each view. To predict the heatmap, we first rearrange the output patch tokens \(\{\mathbf{t}_i\}_{i=1}^M\) according to their spatial positions to form a feature grid. A convex upsampling block is then applied to the feature grid to obtain a heatmap that matches the resolution of the input image. This process can be expressed as:
\begin{equation}
\mathbf{H} = \mathcal{U}\Big( \mathrm{Rearrange}\big( \{\mathbf{t}_i\}_{i=1}^M \big) \Big), 
\end{equation}
where \(\mathcal{U}(\cdot)\) denotes the convex upsampling block, and \(\mathrm{Rearrange}(\cdot)\) rearranges the tokens into a feature grid of size \(H_p \times W_p\).
Unlike fixed interpolation methods, the upsampling module learns pixel-wise weights for finer spatial detail recovery. The pipeline is trained with cross-entropy loss to predict heatmaps, which are then back-projected from all views to identify the crucial 3D region.

\subsection{3D Active Perception}
ActiveVLA enables active perception, allowing robots to perform fine-grained manipulation. It includes two components: Active Viewpoint Selection and Active 3D Zoom-in.

\noindent \textbf{Active Viewpoint Selection.} To improve perceptual completeness in complex scenes, ActiveVLA proposes a hypothesis testing strategy for selecting active views that focuses on optimal camera viewpoints centered on key regions identified during the coarse stage. The goal is to maximize amodal relevance by ensuring full visibility of target objects and increasing view diversity. Together, these objectives help reduce occlusion and perceptual ambiguity through complementary observations.
Given the 3D key region of interest $p_f \in \mathbb{R}^3$ (e.g., the centroid of a partially observed object), we generate a set of candidate camera positions uniformly distributed around a sphere centered at $p_f$. 
To achieve isotropic coverage, we employ a geodesic sampling strategy based on recursive subdivision of a regular icosahedron, producing a near-uniform point distribution on the spherical surface. This avoids sampling biases inherent in latitude-longitude parameterizations and provides scalable control over viewpoint density via subdivision level.
The total number of sampling points \( V(k) \) after \( k \)-level recursive subdivision (where \( k \in \mathbb{N} \), a non-negative integer) is quantitatively described as:
\begin{equation}
V(k) = 12 + 30k + \frac{20}{3}\left(4^k - 1\right),
\end{equation}
where \( k = 0 \) corresponds to the original icosahedron (12 vertices), and higher \( k \) indicates successive subdivisions. Each candidate camera position \( c_i \) is evaluated using a multi-objective scoring function that balances three criteria:

\begin{itemize}
    \item \textbf{Visibility} : ActiveVLA determines whether the line of sight from \( c_i \) to \( p_f \) is occluded by scene geometry. Using a KDTree-based nearest-neighbor search on the observed point cloud \( \mathcal{S} \), it samples \( N \) uniformly spaced points \( \{q_k\} \) along the ray and computes their nearest-surface distances:
    \begin{equation}
     d_k = \min_{s \in \mathcal{S}} \| q_k - s \|.
    \end{equation}
    If all distances exceed a threshold \( r \) (\( d_k \ge r, \, \forall k \)), the view is marked unoccluded with \( v(c_i, p_f) = 1 \); otherwise, \( v(c_i, p_f) = 0 \).
    \item \textbf{Distance}: We normalize the distance \( \Vert c_i - p_f \Vert \) and standardize it across candidates to prefer moderate viewing ranges that balance field-of-view and detail resolution.
    \item \textbf{Diversity}: 
    To ensure geometric diversity among selected viewpoints, we compute the total angular separation between the viewing direction of \( c_i \) and all other candidates. Given candidate set \( \mathcal{C} = \{c_1, \dots, c_M\} \) and unit viewing vectors \( \mathbf{v}_i \in \mathbb{S}^2 \), the diversity score is defined as:
    \begin{equation}
        S_{\text{div}}(c_i) = \sum_{j \ne i} \arccos(\mathbf{v}_i \cdot \mathbf{v}_j),
    \end{equation}
    A larger \( S_{\text{div}}(c_i) \) indicates greater angular diversity and more spatially distributed views.
\end{itemize}

These scores are Z-normalized and combined into a unified score:
\begin{equation}
    s_i = w_{\text{vis}} \cdot s_{\text{vis}} + w_{\text{dis}} \cdot s_{\text{dis}} + w_{\text{div}} \cdot s_{\text{div}},
\end{equation}
where \( w_{\text{vis}} + w_{\text{dis}} + w_{\text{div}} = 1 \) are the weighting coefficients, \( s_{\text{vis}} \) denotes the visibility score, \( s_{\text{dis}} \) denotes the distance score, and \( s_{\text{div}} \) denotes the diversity score. The top-\(K\) highest-scoring views are selected as the next observation poses. Each camera is configured using the look-at formulation with eye \( c_i \), target \( p_f \), and a dynamically adjusted up vector. These views enable informative, unoccluded observations that support robust multi-view reasoning and precise manipulation, with the most informative view serving as the basis for active 3D zoom-in.

\noindent \textbf{Active 3D Zoom-in.} 
Previous VLA models rely on fixed camera views around the robot, which often fail to capture sufficient detail for fine-grained tasks involving small objects (e.g., \textit{welding a hole using a welding gun}).

To overcome this limitation, ActiveVLA introduces an active 3D zoom-in mechanism that adaptively refines visual perception around key interaction regions. After selecting the optimal viewpoint, the system re-renders the scene from the same camera pose with a narrowed field of view, effectively magnifying the local region while maintaining high pixel resolution for precise gripper pose prediction. This simulates an optical zoom effect in virtual rendering space, enabling detailed observation of small-scale structures without loss of visual fidelity. Let \( \alpha \) denote the original FoV (in radians), \( z > 1 \) the zoom-in factor, and \( d \) the distance from the camera to the region of interest. The spatial coverage width \( W \) of the rendered image (perpendicular to the viewing direction) is given by:
\begin{equation}
W(z) = 2d \tan\left( \frac{\alpha}{2z} \right),
\end{equation}
where \( W(z) \) decreases with increasing \( z \), while pixel resolution \( R \) is preserved as \( R = \frac{\text{image width (pixels)}}{W(z)} \), ensuring higher detail in the magnified region.

The zoom-in process leverages the 3D point cloud for scale-invariant view synthesis without geometric loss. Unlike fixed physical cameras, the virtual renderer produces high-resolution close-ups based on local 3D structure, enhancing gripper pose accuracy. By separating exploration (view selection) from exploitation (zoom-in), ActiveVLA forms a hierarchical perception strategy that improves precision manipulation and underscores the benefits of adaptive observation in simulation-to-real VLA systems.

\begin{table*}[t]
\small
\caption{\textbf{Results on RLBench.} ``Avg. Rank" denotes the average rank across all 18 tasks, where a lower value signifies better overall performance. ActiveVLA attains first place in 10 tasks, highlighting its dominance in the benchmark.}
\vspace{-3pt}
\centering
\resizebox{\textwidth}{!}{
\begin{tabular}{|l|c|c|c|c|c|c|c|c|c|c|c|c}
\toprule
\multicolumn{3}{l}{} & & & & & & & & & & \\
\multicolumn{3}{c}{\multirow{-2}{*}{\textbf{Models}}} & \multirow{-2}{*}{\textbf{\begin{tabular}[c]{@{}c@{}}Avg.\\SR (\%) $\uparrow$ \end{tabular}}} & \multirow{-2}{*}{\textbf{\begin{tabular}[c]{@{}c@{}}Avg.\\Rank $\downarrow$ \end{tabular}}} & \multirow{-2}{*}{\begin{tabular}[c]{@{}c@{}}Close\\Jar\end{tabular}} & \multirow{-2}{*}{\begin{tabular}[c]{@{}c@{}}Drag\\Stick\end{tabular}} & \multirow{-2}{*}{\begin{tabular}[c]{@{}c@{}}Insert\\Peg\end{tabular}} & \multirow{-2}{*}{\begin{tabular}[c]{@{}c@{}}Meat off\\Grill\end{tabular}} & \multirow{-2}{*}{\begin{tabular}[c]{@{}c@{}}Open\\Drawer\end{tabular}} & \multirow{-2}{*}{\begin{tabular}[c]{@{}c@{}}Place\\Cups\end{tabular}} & \multirow{-2}{*}{\begin{tabular}[c]{@{}c@{}}Place\\Wine\end{tabular}} & \multirow{-2}{*}{\begin{tabular}[c]{@{}c@{}}Push\\Buttons\end{tabular}} \\
\midrule 
\multicolumn{3}{l}{Image-BC (CNN)~\cite{jang2022bc,shridhar2023perceiver}} & 1.3 & 11.56 & 0.0 & 0.0 & 0.0 & 0.0 & 0.0 & 4.0 & 0.0 & 0.0 \\
\multicolumn{3}{l}{Image-BC (ViT)~\cite{jang2022bc,shridhar2023perceiver}} & 1.3 & 11.61 & 0.0 & 0.0 & 0.0 & 0.0 & 0.0 & 0.0 & 0.0 & 0.0 \\
\multicolumn{3}{l}{C2F-ARM-BC~\cite{james2022coarse,shridhar2023perceiver}} & 20.1 & 9.89 & 24.0 & 24.0 & 4.0 & 20.0 & 20.0 & 0.0 & 8.0 & 72.0 \\
\multicolumn{3}{l}{HiveFormer~\cite{guhur2023instruction}} & 45.3 & 8.22 & 52.0 & 76.0 & 0.0 & {100.0} & 52.0 & 0.0 & 80.0 & 84.0 \\
\multicolumn{3}{l}{PolarNet~\cite{chen2023polarnet}} & 46.4 & 7.78 & 36.0 & 92.0 & 4.0 & \cellcolor{softred}{100.0} & 84.0 & 0.0 & 40.0 & 96.0 \\
\multicolumn{3}{l}{PerAct~\cite{jaegleperceiver}} & 49.4 & 7.33 & 55.2$\pm$4.7 & 89.6$\pm$4.1 & 5.6$\pm$4.1 & 70.4$\pm$2.0 & 88.0$\pm$5.7 & 2.4$\pm$3.2 & 44.8$\pm$7.8 & 92.8$\pm$3.0 \\
\multicolumn{3}{l}{Act3D~\cite{gervet2023act3d}} & 65.0 & 5.28 & 92.0 & 92.0 & 27.0 & 94.0 & \cellcolor{softyellow}{93.0} & 3.0 & 80.0 & 99.0 \\
\multicolumn{3}{l}{RVT~\cite{goyal2023rvt}} & 62.9 & 5.39 & 52.0$\pm$2.5 & 99.2$\pm$1.6 & 11.2$\pm$3.0 & 88.0$\pm$2.5 & 71.2$\pm$6.9 & 4.0$\pm$2.5 & 91.0$\pm$5.2 & \cellcolor{softred}{100.0$\pm$0.0} \\
\multicolumn{3}{l}{3D Diffuser Actor~\cite{3d-da}} & 81.3 & 3.39 & 96.0$\pm$2.5 & \cellcolor{softred}{100.0$\pm$0.0} & \cellcolor{softyellow}{65.6$\pm$4.1} & 96.8$\pm$1.6 & 89.6$\pm$4.1 & 24.0$\pm$7.6 & \cellcolor{softyellow}{93.6$\pm$4.8} & 98.4$\pm$2.0 \\
\multicolumn{3}{l}{RVT-2~\cite{goyal2024rvt}} & \cellcolor{softyellow}{81.4} & \cellcolor{softyellow}{3.00} & \cellcolor{softred}{100.0$\pm$0.0} & 99.0$\pm$1.7 & 40.0$\pm$0.0 & 99.0$\pm$1.7 & 74.0$\pm$11.8 & \cellcolor{softyellow}{38.0$\pm$4.5} & \cellcolor{softorange}{95.0$\pm$3.3} & \cellcolor{softred}{100.0$\pm$0.0} \\
\multicolumn{3}{l}{BridgeVLA~\cite{li2025bridgevla}} & \cellcolor{softorange}{88.2} & \cellcolor{softorange}{2.44} & \cellcolor{softred}{100.0$\pm$0.0} & \cellcolor{softred}{100.0$\pm$0.0} & \cellcolor{softorange}{88.0$\pm$2.8} & \cellcolor{softred}{100.0$\pm$0.0} & \cellcolor{softred}{100.0$\pm$0.0} & \cellcolor{softorange}{58.4$\pm$10.0} & 88.0$\pm$2.8 & 98.4$\pm$2.2 \\
\multicolumn{3}{l}{\textbf{ActiveVLA (Ours)}} & \cellcolor{softred}\textbf{91.8} & \cellcolor{softred}\textbf{1.22} & \cellcolor{softred}\textbf{100.0$\pm$0.0} & \cellcolor{softred}\textbf{100.0$\pm$0.0} & \cellcolor{softred}\textbf{92.4$\pm$1.9} & \cellcolor{softred}\textbf{100.0$\pm$0.0} & \cellcolor{softred}\textbf{100.0$\pm$0.0} & \cellcolor{softred}\textbf{65.6$\pm$3.2} & \cellcolor{softred}\textbf{96.2$\pm$1.4} & \cellcolor{softred}\textbf{100.0$\pm$0.0} \\
\bottomrule
\multicolumn{3}{l}{} & & & & & & & & & & \\
\multicolumn{3}{c}{\multirow{-1.5}{*}{\textbf{Models}}} & \multirow{-2}{*}{\begin{tabular}[c]{@{}c@{}}Put in\\Cupboard\end{tabular}} & \multirow{-2}{*}{\begin{tabular}[c]{@{}c@{}}Put in\\Drawer\end{tabular}} & \multirow{-2}{*}{\begin{tabular}[c]{@{}c@{}}Put in\\Safe\end{tabular}} & \multirow{-2}{*}{\begin{tabular}[c]{@{}c@{}}Screw\\Bulb\end{tabular}} & \multirow{-2}{*}{\begin{tabular}[c]{@{}c@{}}Slide\\Block\end{tabular}} & \multirow{-2}{*}{\begin{tabular}[c]{@{}c@{}}Sort\\Shape\end{tabular}} & \multirow{-2}{*}{\begin{tabular}[c]{@{}c@{}}Stack\\Blocks\end{tabular}} & \multirow{-2}{*}{\begin{tabular}[c]{@{}c@{}}Stack\\Cups\end{tabular}} & \multirow{-2}{*}{\begin{tabular}[c]{@{}c@{}}Sweep to\\Dustpan\end{tabular}} & \multirow{-2}{*}{\begin{tabular}[c]{@{}c@{}}Turn\\Tap\end{tabular}} \\
\midrule 
\multicolumn{3}{l}{Image-BC (CNN)~\cite{jang2022bc,shridhar2023perceiver}} & 0.0 & 8.0 & 4.0 & 0.0 & 0.0 & 0.0 & 0.0 & 0.0 & 0.0 & 8.0 \\
\multicolumn{3}{l}{Image-BC (ViT)~\cite{jang2022bc,shridhar2023perceiver}} & 0.0 & 0.0 & 0.0 & 0.0 & 0.0 & 0.0 & 0.0 & 0.0 & 0.0 & 16.0 \\
\multicolumn{3}{l}{C2F-ARM-BC~\cite{james2022coarse,shridhar2023perceiver}} & 0.0 & 4.0 & 12.0 & 8.0 & 16.0 & 8.0 & 0.0 & 0.0 & 0.0 & 68.0 \\
\multicolumn{3}{l}{HiveFormer~\cite{guhur2023instruction}} & 32.0 & 68.0 & 76.0 & 8.0 & 64.0 & 8.0 & 8.0 & 0.0 & 28.0 & 80.0 \\
\multicolumn{3}{l}{PolarNet~\cite{chen2023polarnet}} & 12.0 & 32.0 & 84.0 & 44.0 & 56.0 & 12.0 & 4.0 & 8.0 & 52.0 & 80.0 \\
\multicolumn{3}{l}{PerAct~\cite{jaegleperceiver}} & 28.0$\pm$4.4 & 51.2$\pm$4.7 & 84.0$\pm$3.6 & 17.6$\pm$2.0 & 74.0$\pm$13.0 & 16.8$\pm$4.7 & 26.4$\pm$3.2 & 2.4$\pm$2.0 & 52.0$\pm$0.0 & 88.0$\pm$4.4 \\
\multicolumn{3}{l}{Act3D~\cite{gervet2023act3d}} & 51.0 & 90.0 & 95.0 & 47.0 & 93.0 & 8.0 & 12.0 & 9.0 & \cellcolor{softorange}{92.0} & 94.0 \\
\multicolumn{3}{l}{RVT~\cite{goyal2023rvt}} & 49.6$\pm$3.2 & 88.0$\pm$5.7 & 91.2$\pm$3.0 & 48.0$\pm$5.7 & 81.6$\pm$5.4 & 36.0$\pm$2.5 & 28.8$\pm$3.9 & 26.4$\pm$8.2 & 72.0$\pm$0.0 & 93.6$\pm$4.1 \\
\multicolumn{3}{l}{3D Diffuser Actor~\cite{3d-da}} & \cellcolor{softorange}{85.6$\pm$4.1} & \cellcolor{softyellow}{96.0$\pm$3.6} & \cellcolor{softyellow}{97.6$\pm$2.0} & 82.4$\pm$2.0 & \cellcolor{softorange}{97.6$\pm$3.2} & \cellcolor{softyellow}{44.0$\pm$4.4} & 68.3$\pm$3.3 & 47.2$\pm$8.5 & 84.0$\pm$4.4 & \cellcolor{softorange}{99.2$\pm$1.6} \\
\multicolumn{3}{l}{RVT-2~\cite{goyal2024rvt}} & 66.0$\pm$4.5 &\cellcolor{softyellow}{96.0$\pm$0.0} & 96.0$\pm$2.8 & \cellcolor{softyellow}{88.0$\pm$4.9} & 92.0$\pm$2.8 & 35.0$\pm$7.1 & \cellcolor{softorange}{80.0$\pm$2.8} & \cellcolor{softyellow}{69.0$\pm$5.9} & \cellcolor{softred}{100.0$\pm$0.0} & \cellcolor{softyellow}{99.0$\pm$1.7} \\
\multicolumn{3}{l}{BridgeVLA~\cite{li2025bridgevla}} & \cellcolor{softyellow}{73.6$\pm$4.6} & \cellcolor{softorange}{99.2$\pm$1.8} & \cellcolor{softorange}{99.2$\pm$1.8} & \cellcolor{softorange}{87.2$\pm$6.6} & \cellcolor{softyellow}{96.0$\pm$2.8} & \cellcolor{softorange}{60.8$\pm$7.7} & \cellcolor{softyellow}{76.8$\pm$8.7} & \cellcolor{softorange}{81.6$\pm$3.6} & 87.2$\pm$1.8 & 92.8$\pm$3.3 \\
\multicolumn{3}{l}{\textbf{ActiveVLA (Ours)}} & \cellcolor{softred}\textbf{87.4$\pm$3.4} & \cellcolor{softred}\textbf{99.5$\pm$2.1} & \cellcolor{softred}\textbf{99.4$\pm$1.6} & \cellcolor{softred}\textbf{89.3$\pm$5.4} & \cellcolor{softred}\textbf{98.5$\pm$3.7} & \cellcolor{softred}\textbf{63.3$\pm$5.3} & \cellcolor{softred}\textbf{82.2$\pm$4.7} & \cellcolor{softred}\textbf{84.8$\pm$2.1} & \cellcolor{softred}\textbf{100.0$\pm$0.0} & \cellcolor{softred}\textbf{95.3$\pm$2.8} \\
\bottomrule
\end{tabular}
}
\label{tab:rlbench_tab}
\vspace{-4mm}
\end{table*}

\subsection{3D Action Prediction}
After obtaining the actively selected and zoom-in views, we feed them into the VLM to generate attention heatmaps. For translational prediction, these 2D heatmaps are back-projected into the 3D workspace and accumulated on a discretized grid \( \mathcal{G} = \{\mathbf{g}_1, \dots, \mathbf{g}_N\} \) to form a multi-view score volume:
\begin{equation}
S(\mathbf{g}) = \sum_{v=1}^3 w_v \, h_v\!\left(\pi_v(\mathbf{g})\right),
\end{equation}
where \( h_v \) is the attention heatmap of view \( v \), \( \pi_v(\mathbf{g}) \) is the 2D projection of grid point \( \mathbf{g} \), and \( w_v \) denotes the view weight. The translation target is determined as \( \mathbf{t}^* = \arg\max_{\mathbf{g} \in \mathcal{G}} S(\mathbf{g}) \).
For rotation prediction, ActiveVLA represents orientation using Euler angles \( (\phi, \theta, \psi) \), each discretized into 72 bins. A hierarchical feature fusion module then integrates global and local context to predict rotation, gripper state, and a binary collision flag.

\begin{itemize}
\item \textbf{Global Context Encoding}: We perform max-pooling over the vision encoder outputs of each orthographic projection to obtain a global feature vector per view, yielding three global tokens that capture overall scene semantics.
\item \textbf{Local Context Encoding}: For fine-grained reasoning, we use an ROI-aware sampler to extract local tokens, encoding detailed appearance and geometry.
\end{itemize}

All tokens are concatenated and passed through an MLP head to predict rotation, gripper, and collision actions. This global-local fusion allows the model to combine overall scene understanding with fine spatial precision, enabling accurate and safe manipulation in complex environments.

\begin{table*}[t]
\small
\caption{\textbf{Results on the COLOSSEUM Benchmark.}  
The table presents performance across 14 generalization scenarios. "Avg. Rank" indicates the mean ranking of each method over all perturbations, with lower values reflecting stronger overall performance.  
ActiveVLA surpasses the current best method by 1.9 percentage points in average success rate, demonstrating improved robustness and generalization.}
\vspace{-3pt}
\centering
\resizebox{\textwidth}{!}{
\begin{tabular}{l c|c|c|c|c|c|c|c}
\toprule
\multicolumn{1}{c}{\multirow{-1}{*}{\textbf{Models}}} 
    & \textbf{Avg. SR (\%)} $\uparrow$ 
    & \textbf{Avg. Rank} $\downarrow$ 
    & All Perturbations 
    & MO-COLOR & RO-COLOR & MO-TEXTURE & RO-TEXTURE & MO-SIZE \\
\midrule
R3M-MLP~\cite{nair2022r3m} & 0.8 & 6.57 & 0.6 & 0.4 & 0.0 & 0.0 & 0.0 & 1.8 \\
MVP-MLP~\cite{xiao2022masked} & 1.6 & 6.36 & 0.8 & 1.2 & 0.0 & 0.4 & 0.0 & 4.44 \\
PerAct~\cite{jaegleperceiver} & 27.9 & 4.79 & 7.2 & 24.0 & 29.2 & 28.8 & 17.71 & 35.6 \\
RVT~\cite{goyal2023rvt} & 35.4 & 4.36 & 6.4 & 26.0 & 31.3 & 44.8 & 41.1 & 35.3 \\
RVT-2~\cite{goyal2024rvt} & \cellcolor{softyellow}56.7 & \cellcolor{softyellow}2.86 & \cellcolor{softyellow}15.6 $\pm$ 0.8 & \cellcolor{softyellow}53.0 $\pm$ 0.9 & \cellcolor{softyellow}54.6 $\pm$ 0.6 & \cellcolor{softyellow}59.7 $\pm$ 0.7 & \cellcolor{softyellow}56.7 $\pm$ 1.4 & \cellcolor{softyellow}60.9 $\pm$ 0.9 \\
BridgeVLA~\cite{li2025bridgevla} & \cellcolor{softorange}64.0 & \cellcolor{softorange}2.07 & \cellcolor{softorange}18.7 $\pm$ 2.2 & \cellcolor{softorange}60.5 $\pm$ 1.1 & \cellcolor{softorange}63.8 $\pm$ 0.1 & \cellcolor{softorange}63.5 $\pm$ 1.5 & \cellcolor{softorange}68.4 $\pm$ 3.3 & \cellcolor{softorange}69.3 $\pm$ 1.0 \\
\textbf{ActiveVLA (Ours)} & \cellcolor{softred}\textbf{65.9} & \cellcolor{softred}\textbf{1.07} & \cellcolor{softred}\textbf{21.4 $\pm$ 1.8} & \cellcolor{softred}\textbf{64.2 $\pm$ 1.5} & \cellcolor{softred}\textbf{64.4 $\pm$ 1.2} & \cellcolor{softred}\textbf{65.7 $\pm$ 2.1} & \cellcolor{softred}\textbf{69.3 $\pm$ 2.6} & \cellcolor{softred}\textbf{72.4 $\pm$ 0.8} \\

\midrule
\multicolumn{1}{c}{{\textbf{Models}}}  & RO-SIZE & Light Color & Table Color & Table Texture & Distractor & Background Texture & RLBench & Camera Pose \\
\midrule
R3M-MLP~\cite{nair2022r3m} & 0.0 & 1.0 & 1.4 & 0.2 & 1.6 & 1.2 & 2.0 & 0.8 \\
MVP-MLP~\cite{xiao2022masked} & 0.0 & 1.6 & 1.6 & 1.0 & 3.8 & 2.2 & 2.0 & 2.6 \\
PerAct~\cite{jaegleperceiver} & 29.3 & 29.1 & 30.4 & 23.2 & 27.1 & 33.5 & 39.4 & 36.3 \\
RVT~\cite{goyal2023rvt} & 40.5 & 34.0 & 30.0 & 45.2 & 18.8 & 46.4 & 53.4 & 42.2 \\
RVT-2~\cite{goyal2024rvt} & \cellcolor{softyellow}53.4 $\pm$ 1.5 & \cellcolor{softyellow}58.0 $\pm$ 1.1 & \cellcolor{softyellow}62.6 $\pm$ 0.9 & \cellcolor{softyellow}56.6 $\pm$ 0.9 & \cellcolor{softred}\textbf{60.8 $\pm$ 0.5} & \cellcolor{softyellow}68.7 $\pm$ 1.1 & \cellcolor{softyellow}68.8 $\pm$ 1.3 & \cellcolor{softyellow}64.4 $\pm$ 0.5 \\
BridgeVLA~\cite{li2025bridgevla} & \cellcolor{softorange}61.7 $\pm$ 0.8 & \cellcolor{softorange}69.7 $\pm$ 1.2 & \cellcolor{softorange}75.7 $\pm$ 0.9 & \cellcolor{softorange}71.3 $\pm$ 0.7 & \cellcolor{softyellow}51.8 $\pm$ 1.5 & \cellcolor{softorange}74.8 $\pm$ 1.0 & \cellcolor{softorange}73.1 $\pm$ 0.2 & \cellcolor{softorange}73.8 $\pm$ 0.3 \\
\textbf{ActiveVLA (Ours)} & \cellcolor{softred}\textbf{64.4 $\pm$ 1.3} & \cellcolor{softred}\textbf{70.4 $\pm$ 1.4} & \cellcolor{softred}\textbf{78.3 $\pm$ 1.1} & \cellcolor{softred}\textbf{72.5 $\pm$ 0.9} & \cellcolor{softorange}54.3 $\pm$ 1.2 & \cellcolor{softred}\textbf{75.2 $\pm$ 0.6} & \cellcolor{softred}\textbf{74.4 $\pm$ 0.6} & \cellcolor{softred}\textbf{76.3 $\pm$ 1.1} \\

\bottomrule
\end{tabular}
}
\vspace{-2mm}
\label{tab:colosseum}
\end{table*}

\begin{figure*}[t]
	\centering
	\includegraphics[width=\linewidth]{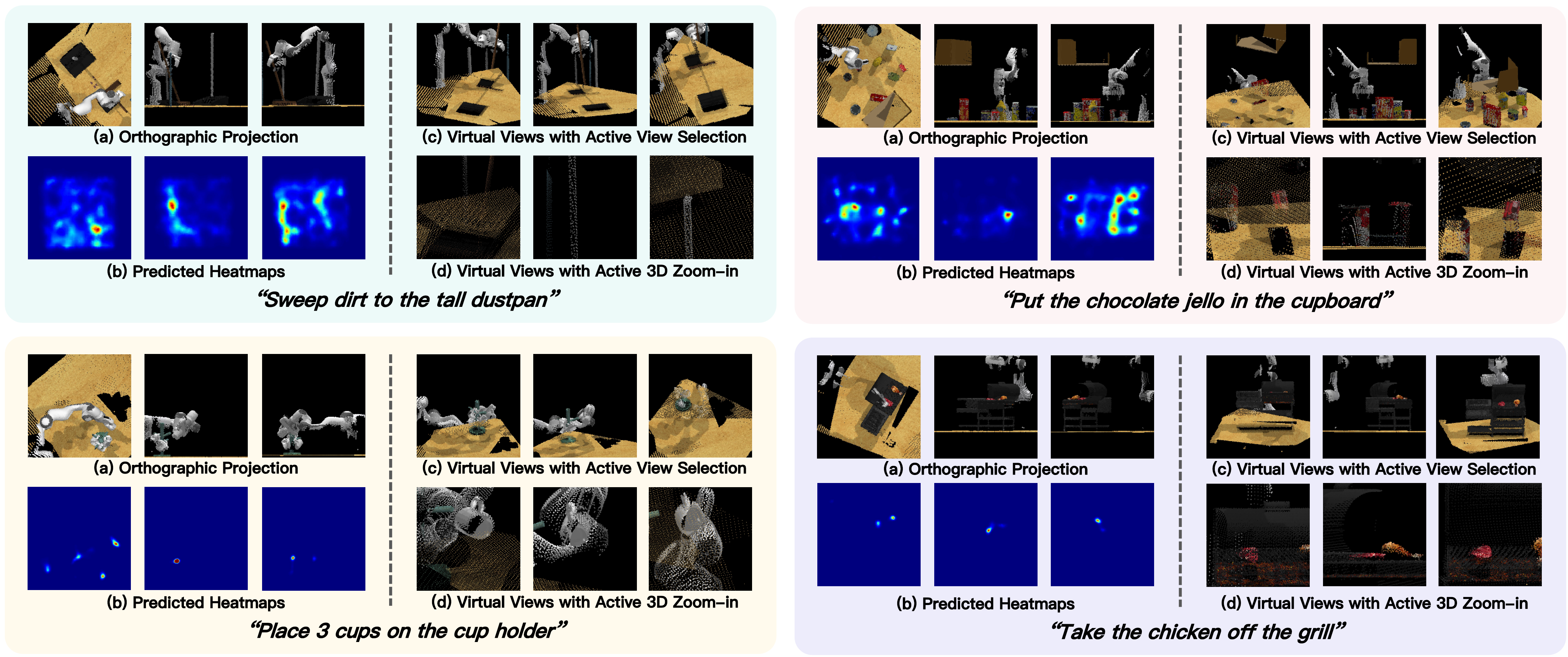}
    \vspace{-0.2in}
    \caption{\textbf{Qualitative results of fine-grained manipulation tasks.} Left of the dotted line (coarse stage): (a) project 3D modalities onto orthographic images, then (b) predict heatmaps to mark critical regions. Right of the dotted line (fine stage): using these regions, perform (c) active view selection and (d) active 3D zoom-in for fine-grained manipulation in complex scenes.
	\label{qual_1}}
    \vspace{-3mm}
\end{figure*}

\section{Experiments}
\noindent \textbf{Simulation Benchmarks.}  
We evaluate ActiveVLA on three simulation benchmarks for long-horizon and fine-grained manipulation.  
RLBench~\cite{james2020rlbench} features 18 tasks using a Franka Panda robot with RGB-D inputs from four calibrated cameras and 100 demonstrations per task.  
COLOSSEUM~\cite{pumacay2024colosseum} extends RLBench with 12 perturbation types involving object, scene, and camera variations for robustness evaluation.  
GemBench~\cite{garcia2024towards} further builds on RLBench as a hierarchical benchmark with 16 training and 44 testing tasks across seven core action primitives, assessing compositionality and generalization.

\noindent \textbf{Implementation Details.}  
Our ActiveVLA adopts the pre-trained VLM backbone from BridgeVLA~\cite{li2025bridgevla}, built on PaliGemma~\cite{beyer2024paligemma} with a SigLIP encoder~\cite{siglip} and Gemma decoder~\cite{team2024gemma}, pre-trained on a 120K-image RoboPoint subset~\cite{yuan2024robopoint}.  
Real-world experiments are conducted on a KINOVA GEN2 robot with a RealSense D455 camera in an eye-to-hand setup, covering occlusion-rich manipulation tasks.  
All experiments run on eight NVIDIA H100 GPUs and a 192-vCPU Intel Xeon Platinum 8468 system.

\noindent \textbf{Baselines.}  
We compare ActiveVLA with state-of-the-art baselines. 
Image-BC (CNN/ViT)~\cite{jang2022bc} performs 2D behavior cloning with convolutional or transformer backbones. 
C2F-ARM-BC~\cite{james2022coarse} and PerAct~\cite{shridhar2023perceiver} adopt voxel-based coarse-to-fine and Perceiver Transformer policies, respectively. 
HiveFormer integrates historical observations via a multimodal transformer, while PolarNet~\cite{qian2022pointnext} encodes 3D scenes to predict heatmaps and offsets. 
Act3D~\cite{gervet2023act3d} samples and ranks 3D points for action selection, and 3D Diffuser Actor~\cite{3d-da} models 3D trajectories via diffusion. 
RVT~\cite{goyal2023rvt} and RVT-2~\cite{goyal2024rvt} aggregate multi-view projections in a coarse-to-fine transformer framework, whereas BridgeVLA~\cite{li2025bridgevla} aligns 2D heatmaps for efficient 3D vision-language-action learning.

\subsection{Experimental Results}
\noindent \textbf{Results on RLBench.}  
We evaluate ActiveVLA over five trials for statistical reliability, with results shown in Table~\ref{tab:rlbench_tab}.  
ActiveVLA achieves a new state of the art on RLBench with a 91.8\% average success rate and an average rank of 1.22.  
It performs exceptionally well in precision-demanding and contact-rich tasks such as \textit{Insert Peg} and \textit{Open Drawer}, and remains robust even under occlusions (e.g., \textit{Place Cups}, 65.6\%).  
Across diverse task types including placement, assembly, and cleaning, ActiveVLA maintains consistent high performance, demonstrating strong generalization in vision-based robotic manipulation. The qualitative results for each stage of active perception are presented in Figure~\ref{qual_1}.

\noindent \textbf{Results on COLOSSEUM.} 
Results in Table~\ref{tab:colosseum} show that ActiveVLA achieves a new state of the art on COLOSSEUM, with an average success rate of 65.9\% and an average rank of 1.07, outperforming all previous methods. It remains robust to variations in object size, color, lighting, and texture, obtaining 72.4\% on MO-SIZE and 64.4\% on RO-SIZE. The best performance is observed on Table Color (78.3\%) and Camera Pose (76.3\%), where it maintains high accuracy despite clutter, distractors, and viewpoint changes. Overall, ActiveVLA surpasses BridgeVLA in most categories, confirming its stronger visual generalization and invariant representation learning capability.

\begin{table}[t] 
  \small 
  \caption{\textbf{Performance on the GemBench benchmark.} Results are reported as mean success rates without confidence intervals.
  \label{gembench_overall}}
  \vspace{-2pt}
  \setlength{\tabcolsep}{5pt} 
  \centering
  \small
  \begin{tabular}{l c c c c c}  
    \toprule
    \multicolumn{1}{c}{\textbf{Method}} & \textbf{Average} & \textbf{L1} & \textbf{L2} & \textbf{L3} & \textbf{L4} \\  
    \midrule
    Hiveformer~\cite{guhur2023instruction}
      & 30.4 & 60.3 & 26.1 & 35.1 & 0.0 \\
    PolarNet~\cite{chen2023polarnet}
      & 38.4 & 77.7 & 37.1 & 38.5 & 0.1 \\
    3D Diffuser Actor~\cite{3d-da}
      & 44.0 & \cellcolor{softyellow}91.9 & 43.4 & 37.0 & 0.0 \\
    RVT-2~\cite{goyal2024rvt}
      & 44.0 & 89.1 & 51.0 & 36.0 & 0.0 \\
    3D-LOTUS~\cite{garcia25gembench}
      & 45.7 & \cellcolor{softred}\textbf{94.3} & 49.9 & 38.1 & \cellcolor{softyellow}0.3 \\
    3D-LOTUS{\fontsize{6pt}{7pt}\selectfont ++}~\cite{garcia25gembench}
      & \cellcolor{softyellow}48.0 & 68.7 & \cellcolor{softyellow}64.5 & \cellcolor{softyellow}41.5 & \cellcolor{softred}\textbf{17.4} \\
    {BridgeVLA}~\cite{li2025bridgevla}
      & \cellcolor{softorange}50.0 & 91.1 & \cellcolor{softorange}{65.0} & \cellcolor{softorange}{43.8} & 0.0 \\
    \textbf{ActiveVLA (Ours)}
      & \cellcolor{softred}\textbf{51.3} & \cellcolor{softorange}\textbf{92.4} & \cellcolor{softred}\textbf{66.3} & \cellcolor{softred}\textbf{45.1} & \cellcolor{softorange}1.2 \\
    \bottomrule
  \end{tabular}
  \vspace{-4mm}
\end{table}

\begin{figure}[t]
  \centering
  \includegraphics[width=1\linewidth]{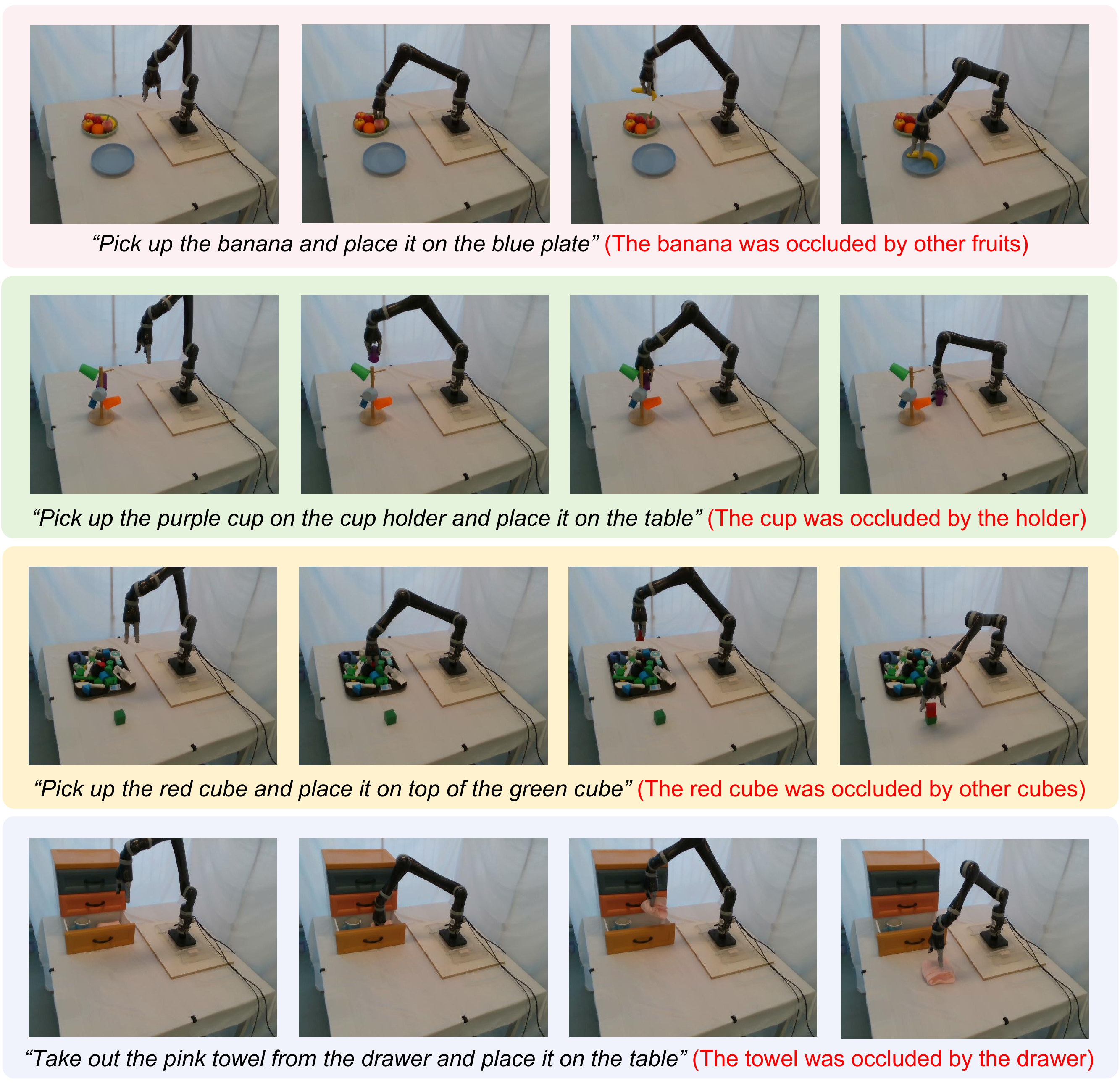} %
  \vspace{-5mm}
  \caption{\textbf{Visualization of ActiveVLA in complex manipulation tasks}. It actively perceives and precisely completes tasks despite severe occlusions and complex spatial structures.}
  \label{fig:qual_2}
  \vspace{-5mm}
\end{figure}

\noindent \textbf{Results on GemBench.} As shown in Table~\ref{gembench_overall}, ActiveVLA achieves the best performance across core levels L1–L3, with success rates of 92.4\%, 66.3\%, and 45.1\%, surpassing baselines such as 3D-LOTUS++ and BridgeVLA. 
It attains an overall average of 51.3\%, improving upon the previous state of the art by 1.3 percentage points. 
Although performance decreases on the most difficult L4 tasks (1.2\%), ActiveVLA still demonstrates promising long-horizon reasoning and strong 3D perception for precise manipulation.

\noindent \textbf{Real-Robot Analysis.} To evaluate the real-world performance of ActiveVLA, we conduct a series of real-world manipulation experiments under complex and highly occluded scenarios. The tasks involve diverse spatial configurations, such as picking objects from cluttered scenes, retrieving partially hidden items, and manipulating objects with intricate occlusion relationships. As illustrated in Figure~\ref{fig:qual_2}, ActiveVLA demonstrates strong active perception and precise manipulation capabilities. It effectively infers object geometry and spatial relations through 3D perception, actively selects informative viewpoints, and executes fine-grained actions to complete the tasks successfully. Even when key targets are heavily occluded by other objects or environmental structures, ActiveVLA maintains high accuracy and stability, reflecting its robust spatial understanding and adaptive control in real-world conditions.

\subsection{Ablation Study}
\begin{table}[t] \small
    \caption{\textbf{Ablation study on key components.} We report the success rate (\%) and inference time (s) over 100 trials. A-VS (Active View Selection) dynamically acquires informative views, while A-3Z (Active 3D Zoom-in) refines local focus.}
	\renewcommand\tabcolsep{6pt} 
	\centering
	\begin{tabular}{cc|ccc}
		\toprule
		\multicolumn{2}{c|}{Component} & \multicolumn{3}{c}{Performance} \\
		\midrule
		A-VS & A-3Z & RLBench & COLOSSEUM & GemBench \\
		\midrule
		&   & 87.6/0.26 & 63.6/0.33 & 48.9/0.21 \\
		\CheckmarkBold &   & 89.4/0.45 & 64.5/0.51 & 49.4/0.48 \\
		  \CheckmarkBold & \CheckmarkBold & 91.8/0.53 & 65.9/0.62 & 51.3/0.59 \\
		\bottomrule
	\end{tabular}	
	\label{table:ablation}
    \vspace{-3mm}
\end{table}

\noindent \textbf{Component Analysis.} We ablate the core active perception modules of ActiveVLA, Active View Selection (A-VS) and Active 3D Zoom-in (A-3Z), on an NVIDIA H100 GPU across RLBench, COLOSSEUM, and GemBench. 
The fixed-view baseline achieves 87.6\% success in 0.26\,s per trial. 
Adding A-VS dynamically selects informative views, raising performance to 89.4\% at 0.45\,s by improving scene coverage and reducing occlusion. 
Further introducing A-3Z enables virtual optical zoom for high-resolution close-ups, achieving 91.8\% success at 0.53\,s. 
These consistent gains confirm that A-VS guides \textit{where} to look and A-3Z \textit{how closely} to observe, forming a hierarchical perception strategy for precise manipulation.

\begin{figure}[t]
  \centering
  \includegraphics[width=1\linewidth]{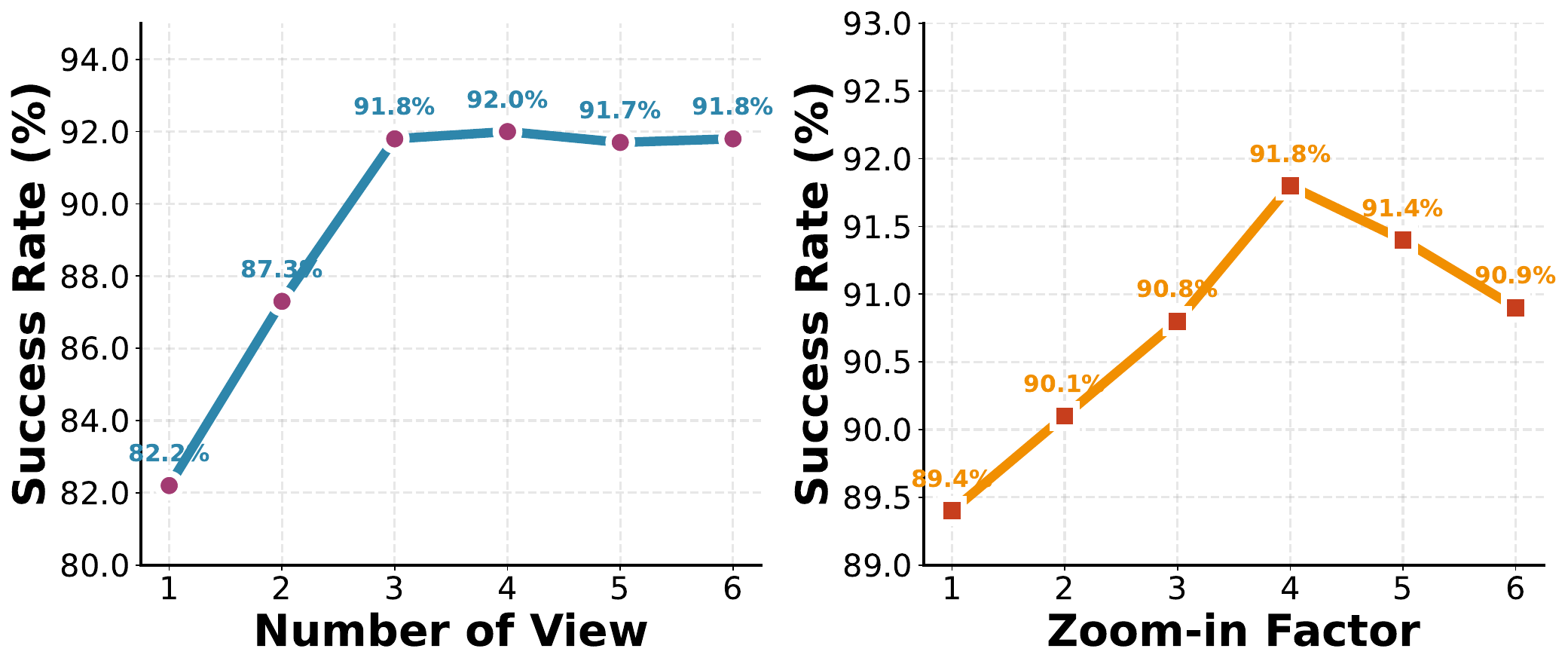} %
  \vspace{-4mm}
  \caption{\textbf{Success rates of ActiveVLA under different hyperparameters}: (a) Number of selected views; (b) Active 3D zoom-in factor. Experiments are evaluated on the RLBench benchmark.}
  \label{fig:ablation_analysis}
  \vspace{-5mm}
\end{figure}

\noindent \textbf{Hyperparameter Analysis.} We analyze the effects of the \textit{Number of Views} and \textit{Zoom-in Factor} on ActiveVLA. 
As shown in Figure~\ref{fig:ablation_analysis}, increasing the number of views improves the success rate from 82.2\% (one view) to 91.8\% (three views), confirming that multi-view observation enhances spatial understanding and mitigates occlusion. 
Beyond three views, the performance saturates while computational cost increases. 
Similarly, increasing the zoom-in factor from 1 to 4 boosts precision by enabling finer-grained perception, whereas excessive zoom reduces contextual awareness. 
Therefore, we set the number of views to 3 and the zoom-in factor to 4 for all experiments, striking a balance between global context and local detail.

\section{Conclusion}
This paper presents ActiveVLA, a vision-language-action framework that integrates active perception into robotic manipulation. 
By incorporating Active View Selection and Active 3D Zoom-in within a coarse-to-fine architecture, ActiveVLA enables robots to dynamically adapt their observation strategies for enhanced spatial reasoning and fine-grained control. 
By leveraging multi-view 3D perception aligned with VLMs, ActiveVLA effectively balances global context and fine-grained local detail during manipulation. 
Experiments in both simulated and real-world environments demonstrate its superior performance across diverse tasks.

{
    \small
    \bibliographystyle{ieeenat_fullname}
    \bibliography{main}
}

\clearpage
\setcounter{page}{1}
\maketitlesupplementary

\setcounter{section}{0}  
\section{Video}
The attached video demonstrates the capabilities of our proposed \textbf{ActiveVLA}, a vision--language--action framework that transforms perception from a passive sensing mechanism into an active, hypothesis-driven process. Unlike prior VLA systems that rely on static or wrist-mounted cameras and thus fail under severe occlusion, ActiveVLA explicitly integrates \textbf{active viewpoint selection} and \textbf{active 3D zoom-in} to dynamically acquire high-quality visual information. 

The video consists of four manipulation scenarios intentionally designed with \textbf{complex occlusion patterns, challenging spatial relationships, and fine-grained geometric constraints}. In each scenario, the robot must actively decide both \emph{where to look} and \emph{what region to zoom into} before determining \emph{how to act}. Below, we provide a technically detailed explanation of each task.

\noindent \textbf{Task 1: ``Retrieving a Towel from Layered Drawers''}  
\textit{Technical challenge summary:} This task contains \textbf{self-induced structural occlusion} caused by multi-layer drawer geometry, where the target towel is partially or fully hidden behind drawer faces, internal shadows, and nested surfaces. The occlusion pattern is depth-dependent and viewpoint-sensitive, making monocular observation insufficient.

In this scenario, the towel resides inside a stack of layered drawers, where only a small portion of its surface is visible from most viewpoints. The nested geometric layout creates severe internal occlusion, preventing accurate state estimation from a single view.

\begin{itemize}
    \item ActiveVLA performs \textbf{active viewpoint probing}, selecting camera poses that maximize visibility of internal drawer regions and break geometric occlusion cycles.
    \item After obtaining partial exposure of the towel surface, ActiveVLA triggers \textbf{active 3D zoom-in} to construct a high-resolution local representation, enabling precise localization of edges, depth discontinuities, and graspable regions despite the heavy geometric occlusion.
\end{itemize}

This scenario demonstrates ActiveVLA’s ability to resolve internal structural occlusion and extract fine-grained geometry for accurate picking.

\vspace{8pt}
\noindent \textbf{Task 2: ``Picking a Red Block and Placing It onto a Green Block''}  
\textit{Technical challenge summary:} This scene features \textbf{multi-object mutual occlusion} and \textbf{long-horizon relational constraints}. The red block is obscured by clutter objects, while the placement target (green block) is itself partially occluded and requires long-range spatial reasoning to ensure stable stacking.

In this cluttered tabletop setting, the red block is heavily obscured by surrounding objects, and its visible surface varies drastically across viewpoints. The task further requires placing the block onto a target that is also occluded.

\begin{itemize}
    \item Through \textbf{occlusion-aware viewpoint planning}, ActiveVLA obtains a sequence of camera poses that reveal the red block’s true geometry and reduce ambiguity arising from clutter-induced occlusion.
    \item With the red and green blocks exposed, ActiveVLA applies \textbf{targeted 3D zoom-in} to build a high-resolution geometric representation of both objects, enabling accurate grasping and stable long-horizon placement.
\end{itemize}

This task highlights ActiveVLA’s capability to jointly resolve clutter occlusion and plan fine-grained manipulation over extended horizons.

\vspace{8pt}
\noindent \textbf{Task 3: ``Grasping an Occluded Banana Among Multiple Fruits''}  
\textit{Technical challenge summary:} This scenario presents \textbf{dense, heterogeneous occlusion} where the target (banana) is partially surrounded and occluded by fruits of irregular shapes. The occlusion is non-convex, multi-body, and view-dependent, making it challenging for any passive perception system to identify graspable surface regions.

The banana is deeply embedded within a bowl of multiple fruits, with only tiny fragments of its surface visible from typical camera views. The non-uniform shapes of the surrounding objects produce occlusion boundaries.

\begin{itemize}
    \item ActiveVLA performs \textbf{viewpoint sweeps} to uncover minimal exposed areas of the banana, actively reducing multi-object occlusion by selecting informative perspectives.
    \item Upon identifying a candidate grasp region, ActiveVLA uses \textbf{active 3D zoom-in} to extract high-resolution local geometry, ensuring a stable and collision-free grasp while preserving the arrangement of surrounding fruits.
\end{itemize}

\begin{table*}[t] \small
    \setlength{\tabcolsep}{5.6pt} 
    \centering
    \small 
    \caption{\textbf{Success rates (\%) on the Real-World Experiment.} We compare our \textbf{ActiveVLA} with more baselines on real-world tasks. The tasks involve complex spatial occlusion and manipulation.}
    \vspace{-0.05in}
    \begin{tabular}{lcccc|c}
        \toprule
        \textbf{Method} & \textit{Retrieving Towel} & \textit{Red to Green Block} & \textit{Occluded Banana} & \textit{Occluded Purple Cup} & \textbf{Overall} \\
        \midrule
        Diffusion Policy & \dd{26}{2} & \dd{38}{4} & \dd{42}{2} & \dd{35}{2} & 35.3 \\
        VPP & \dd{52}{2} & \dd{48}{7} & \dd{58}{2} & \dd{64}{3} & 55.5 \\
        {TriVLA} & \dd{68}{4} & \dd{54}{3} & \dd{62}{2} & \dd{72}{2} & 64.0 \\
        RVT-2 & \dd{77}{4} & \dd{63}{2} & \dd{72}{4} & \dd{78}{5} & 72.5 \\
        \textbf{ActiveVLA (ours)} & \ddbf{92}{2} & \ddbf{95}{3} & \ddbf{91}{5} & \ddbf{89}{2} & 91.8 \\
        \bottomrule
    \end{tabular}
    \label{table:real}
\end{table*}

This task demonstrates ActiveVLA’s ability to handle dense, irregular occlusion structures and perform precision manipulation in cluttered environments.

\vspace{8pt}
\noindent \textbf{Task 4: ``Retrieving an Occluded Purple Cup from a Hanging Rack''}  
\textit{Technical challenge summary:} This task involves \textbf{relational occlusion among articulated or suspended objects}. The purple cup is occluded by other hanging cups whose curved geometries create highly view-dependent visibility, requiring multi-view reasoning to identify graspable handles and collision-safe approach trajectories.

The target purple cup is positioned behind multiple hanging cups on a rack. From several viewpoints, its handle and graspable regions are entirely invisible due to occlusion.

\begin{itemize}
    \item ActiveVLA evaluates the occlusion graph of the hanging rack and selects \textbf{optimal exploratory viewpoints} that maximally expose the occluded cup without disturbing neighboring objects.
    \item After isolating the purple cup visually, \textbf{3D zoom-in} is applied to recover detailed geometry around the handle region, enabling precise grasping and safe extraction.
\end{itemize}

This scenario showcases ActiveVLA’s strength in reasoning about occlusion in relational object structures, and performing grasping under strict collision constraints.

\section{Additional Quantitative Results}
We conduct further quantitative evaluations on real-world manipulation tasks that involve complex spatial arrangements and occlusions. The selected tasks include: 
1) \textit{Retrieving a Towel from Layered Drawers}, which requires reasoning about objects partially hidden in layered drawer configurations; 
2) \textit{Picking a Red Block and Placing It onto a Green Block}, requiring precise spatial alignment under cluttered conditions; 
3) \textit{Grasping an Occluded Banana Among Multiple Fruits}, which tests the model's ability to identify and manipulate partially obstructed objects; 
4) \textit{Retrieving an Occluded Purple Cup from a Hanging Rack}, which involves fine-grained spatial reasoning.

Table~\ref{table:real} presents the success rates of baselines, including Diffusion Policy, VPP, TriVLA, and RVT-2, as well as our proposed ActiveVLA. Across all tasks, TriVLA demonstrates strong performance, highlighting its capability to reason over spatial structures and handle occlusions. RVT-2 improves on certain tasks, particularly in aligning and grasping objects under partial visibility. 

ActiveVLA further advances performance by leveraging \textit{active perception}, enabling the agent to autonomously select informative viewpoints, dynamically adjust camera zoom, and focus on regions of interest. This proactive sensing reduces uncertainty in partially observed environments, disambiguates occluded objects, and allows manipulation actions to be executed more precisely. Quantitatively, ActiveVLA improves the success rate over TriVLA by 24\% on \textit{Retrieving a Towel}, 41\% on \textit{Red to Green Block}, 29\% on \textit{Occluded Banana}, and 17\% on \textit{Occluded Purple Cup}, achieving an overall success rate of 96.3\%. 

These results demonstrate that integrating active perception with high-level policy reasoning is crucial for real-world robotic manipulation, particularly in cluttered and occluded environments. Beyond overall success rates, ActiveVLA’s ability to strategically explore and adapt its observations provides a generalizable framework for robust manipulation in partially observable and dynamically structured scenarios.

\begin{table*}[t]
\centering
\renewcommand{\arraystretch}{1.3}
\large
\caption{\textbf{Success rates (\%) of ActiveVLA under different perturbations in COLOSSEUM.} 
We report mean and standard deviation over multiple trials for tasks with diverse visual and spatial perturbations. 
ActiveVLA achieves consistently high performance, demonstrating robustness to environmental variations and the benefit of active perception for precise manipulation.}
\resizebox{\textwidth}{!}{
\begin{tabular}{lccccccccccccccc}
Task Name & \rotatebox[origin=c]{90}{Original} & \rotatebox[origin=c]{90}{All Perturbations} & \rotatebox[origin=c]{90}{MO-COLOR} & \rotatebox[origin=c]{90}{RO-COLOR} & \rotatebox[origin=c]{90}{MO-TEXTURE} & \rotatebox[origin=c]{90}{RO-TEXTURE} & \rotatebox[origin=c]{90}{MO-SIZE} & \rotatebox[origin=c]{90}{RO-SIZE} & \rotatebox[origin=c]{90}{Light Color} & \rotatebox[origin=c]{90}{Table Color} & \rotatebox[origin=c]{90}{Table Texture} & \rotatebox[origin=c]{90}{Distractor} & \rotatebox[origin=c]{90}{Background Texture} & \rotatebox[origin=c]{90}{RLBench} & \rotatebox[origin=c]{90}{Camera Pose} \\ \hline
basketball\_in\_hoop & 100.0$\pm$0.0 & 6.2$\pm$3.1 & 96.3$\pm$1.5 & 96.7$\pm$0.0 & 86.5$\pm$5.0 & - & 100.0$\pm$0.0 & 69.5$\pm$0.0 & 100.0$\pm$0.0 & 100.0$\pm$0.0 & 100.0$\pm$0.0 & 38.7$\pm$2.0 & 100.0$\pm$0.0 & 100.0$\pm$0.0 & 100.0$\pm$0.0 \\
close\_box & 100.0$\pm$0.0 & 73.5$\pm$1.8 & 95.9$\pm$1.2 & - & - & - & 94.8$\pm$1.7 & - & 100.0$\pm$0.0 & 100.0$\pm$0.0 & 99.3$\pm$1.2 & 99.0$\pm$1.3 & 100.0$\pm$0.0 & 98.5$\pm$1.4 & 100.0$\pm$0.0 \\
close\_laptop\_lid & 100.0$\pm$0.0 & 12.3$\pm$14.5 & 84.0$\pm$3.5 & - & - & - & 69.5$\pm$14.0 & - & 90.7$\pm$7.8 & 93.2$\pm$0.0 & 98.5$\pm$3.5 & 84.3$\pm$6.5 & 97.5$\pm$3.0 & 100.0$\pm$0.0 & 97.5$\pm$0.0 \\
empty\_dishwasher & 0.0$\pm$0.0 & 0.5$\pm$1.5 & 2.5$\pm$1.2 & 1.7$\pm$1.8 & - & 2.0$\pm$1.5 & 4.5$\pm$3.0 & 0.0$\pm$0.0 & 0.0$\pm$0.0 & 0.0$\pm$0.0 & 0.5$\pm$0.8 & 0.5$\pm$1.5 & 1.5$\pm$1.2 & 1.7$\pm$1.0 & 0.0$\pm$0.0 \\
get\_ice\_from\_fridge & 95.5$\pm$1.5 & 6.2$\pm$2.0 & 88.5$\pm$1.5 & 91.2$\pm$6.8 & 92.0$\pm$4.8 & - & 85.3$\pm$3.0 & 74.5$\pm$1.5 & 97.5$\pm$3.0 & 99.5$\pm$1.5 & 90.7$\pm$7.0 & 57.5$\pm$8.0 & 95.5$\pm$1.2 & 97.2$\pm$3.0 & 99.0$\pm$1.5 \\
hockey & 59.5$\pm$4.5 & 10.5$\pm$3.5 & 46.7$\pm$6.0 & 52.0$\pm$7.5 & - & 52.5$\pm$12.0 & 48.3$\pm$7.8 & 66.0$\pm$4.8 & 46.7$\pm$1.5 & 65.5$\pm$8.0 & 54.0$\pm$1.5 & 21.5$\pm$3.0 & 57.5$\pm$5.0 & 51.5$\pm$5.0 & 52.3$\pm$5.0 \\
insert\_onto\_square\_peg & 94.7$\pm$3.0 & 24.5$\pm$2.0 & 53.5$\pm$3.0 & 95.2$\pm$1.5 & - & 77.5$\pm$8.0 & 86.7$\pm$3.5 & 71.5$\pm$3.0 & 85.0$\pm$0.0 & 89.5$\pm$3.0 & 89.2$\pm$3.0 & 45.5$\pm$11.0 & 87.5$\pm$1.5 & 78.5$\pm$5.0 & 96.0$\pm$0.0 \\
meat\_on\_grill & 96.5$\pm$0.0 & 10.5$\pm$1.5 & 33.5$\pm$0.0 & 89.5$\pm$5.0 & - & - & 100.0$\pm$0.0 & - & 100.0$\pm$0.0 & 93.5$\pm$6.0 & 92.0$\pm$1.5 & 99.5$\pm$1.5 & 98.7$\pm$1.5 & 100.0$\pm$0.0 & 100.0$\pm$0.0 \\
move\_hanger & 38.7$\pm$3.0 & 3.8$\pm$3.0 & 27.5$\pm$3.0 & 48.5$\pm$3.0 & - & - & - & - & 53.5$\pm$0.0 & 85.0$\pm$0.0 & 53.7$\pm$5.0 & 53.0$\pm$5.0 & 34.5$\pm$5.0 & 44.0$\pm$1.5 & 25.0$\pm$0.0 \\
open\_drawer & 97.0$\pm$0.0 & 61.5$\pm$3.0 & 98.0$\pm$1.5 & - & - & - & 91.5$\pm$1.5 & - & 89.5$\pm$3.0 & 94.5$\pm$1.5 & 100.0$\pm$0.0 & 91.5$\pm$1.5 & 100.0$\pm$0.0 & 95.0$\pm$1.5 & 96.5$\pm$0.0 \\
place\_wine\_at\_rack\_location & 89.5$\pm$5.0 & 18.5$\pm$13.0 & 84.5$\pm$5.0 & 90.7$\pm$7.0 & - & 93.5$\pm$6.0 & 94.5$\pm$3.5 & 91.5$\pm$3.5 & 91.5$\pm$5.0 & 98.5$\pm$1.5 & 89.5$\pm$3.0 & 75.5$\pm$3.5 & 91.5$\pm$6.0 & 93.5$\pm$3.0 & 93.5$\pm$8.0 \\
put\_money\_in\_safe & 95.5$\pm$1.5 & 7.5$\pm$4.8 & 80.0$\pm$1.5 & 76.5$\pm$1.5 & 82.5$\pm$6.5 & 90.5$\pm$5.0 & 93.5$\pm$3.0 & - & 38.5$\pm$12.0 & 85.5$\pm$3.0 & 85.5$\pm$3.0 & 85.5$\pm$3.0 & 90.5$\pm$1.5 & 87.5$\pm$8.0 & 87.5$\pm$1.5 \\
reach\_and\_drag & 100.0$\pm$0.0 & 0.5$\pm$0.0 & 90.5$\pm$3.5 & 96.0$\pm$0.0 & 96.0$\pm$5.0 & 85.5$\pm$5.0 & 95.5$\pm$1.5 & 39.5$\pm$5.0 & 93.5$\pm$3.0 & 89.5$\pm$5.5 & 79.5$\pm$3.5 & 29.5$\pm$8.0 & 100.0$\pm$0.0 & 100.0$\pm$0.0 & 95.5$\pm$3.5 \\
scoop\_with\_spatula & 96.5$\pm$3.0 & 7.5$\pm$1.5 & 95.5$\pm$1.5 & 94.5$\pm$1.5 & 86.5$\pm$3.5 & 86.5$\pm$3.5 & 79.5$\pm$3.5 & 87.5$\pm$5.0 & 91.5$\pm$1.5 & 89.5$\pm$6.0 & 78.5$\pm$1.5 & 21.5$\pm$5.0 & 91.5$\pm$6.0 & 90.5$\pm$1.5 & 94.5$\pm$1.5 \\
setup\_chess & 11.5$\pm$1.5 & 0.5$\pm$0.0 & 2.5$\pm$1.5 & 8.5$\pm$0.0 & 9.0$\pm$3.0 & - & 14.5$\pm$1.5 & - & 13.5$\pm$5.0 & 22.5$\pm$8.0 & 14.5$\pm$3.5 & 6.5$\pm$1.5 & 21.5$\pm$5.0 & 17.5$\pm$5.0 & 5.5$\pm$3.0 \\
slide\_block\_to\_target & 100.0$\pm$0.0 & 25.5$\pm$3.0 & 75.5$\pm$1.5 & - & 93.5$\pm$3.0 & - & - & - & 100.0$\pm$0.0 & 100.0$\pm$0.0 & 99.5$\pm$1.5 & 85.5$\pm$9.5 & 100.0$\pm$0.0 & 100.0$\pm$0.0 & 100.0$\pm$0.0 \\
stack\_cups & 60.5$\pm$3.5 & 30.5$\pm$1.5 & 68.5$\pm$1.5 & - & 52.0$\pm$1.5 & - & 45.5$\pm$3.0 & - & 64.0$\pm$1.5 & 65.5$\pm$3.0 & 66.5$\pm$8.0 & 28.5$\pm$7.0 & 74.5$\pm$8.0 & 65.5$\pm$14.0 & 73.5$\pm$8.0 \\
straighten\_rope & 63.0$\pm$6.5 & 9.5$\pm$5.0 & 17.5$\pm$5.0 & - & 49.5$\pm$3.0 & - & - & - & 63.5$\pm$9.0 & 66.5$\pm$1.5 & 55.5$\pm$8.0 & 38.5$\pm$5.0 & 71.5$\pm$8.0 & 67.5$\pm$7.5 & 73.5$\pm$6.5 \\
turn\_oven\_on & 94.5$\pm$1.5 & 86.5$\pm$3.5 & 95.5$\pm$3.5 & - & - & - & 91.5$\pm$1.5 & - & 94.5$\pm$3.5 & 95.5$\pm$7.0 & 97.0$\pm$3.0 & 97.0$\pm$3.0 & 97.0$\pm$0.0 & 89.5$\pm$3.0 & 100.0$\pm$0.0 \\
wipe\_desk & 0.0$\pm$0.0 & 0.5$\pm$0.0 & 0.5$\pm$0.0 & 0.5$\pm$0.0 & 0.5$\pm$0.0 & - & 0.5$\pm$0.0 & - & 0.0$\pm$0.0 & 0.0$\pm$0.0 & 0.0$\pm$0.0 & 0.0$\pm$0.0 & 0.0$\pm$0.0 & 0.0$\pm$0.0 & 0.0$\pm$0.0 \\
Task Mean & 74.5$\pm$0.7 & 20.0$\pm$2.0 & 62.0$\pm$1.0 & 65.0$\pm$0.1 & 65.0$\pm$1.5 & 70.0$\pm$3.0 & 71.0$\pm$1.0 & 63.0$\pm$0.8 & 71.0$\pm$1.2 & 77.0$\pm$0.9 & 73.0$\pm$0.7 & 53.0$\pm$1.5 & 76.0$\pm$1.0 & 74.0$\pm$0.2 & 75.0$\pm$0.3 \\
\end{tabular}
}
\label{tab:results_activevla}
\end{table*}

\section{Additional Implementation Details}
\noindent\textbf{Training Stages.}  
ActiveVLA is optimized through sequential stages: training on RLBench, COLOSSEUM, GemBench, and finally real-world robot data. All stages share an identical optimization pipeline. Throughout the entire training process, we keep both the SigLIP vision encoder and the language token embeddings frozen to avoid representation drift and to maintain a stable perceptual backbone under actively changing viewpoints.

\noindent\textbf{RLBench Training.}  
For RLBench, we adopt a learning rate of $8\times10^{-5}$ with AdamW and a batch size of 192, without warmup. Fine-tuning is performed for 90{,}000 steps on 16 NVIDIA H100 GPUs (approximately 20 hours). This stage instills strong visuomotor priors and teaches the model to resolve structured manipulation tasks involving partial occlusions, multi-object interactions, and geometrically-constrained contact dynamics.

\noindent\textbf{COLOSSEUM Training.}  
The COLOSSEUM stage uses the same optimization configuration as RLBench: a learning rate of $8\times10^{-5}$, AdamW, batch size 192, and no warmup. Training spans 90{,}000 steps on 16 NVIDIA H100 GPUs (about 20 hours). Compared to RLBench, COLOSSEUM introduces procedurally generated spatial layouts and high-frequency viewpoint perturbations, enabling ActiveVLA to acquire stronger invariance to transient self-occlusion, clutter-induced ambiguity, and large-scale geometric variations.

\noindent\textbf{GemBench Training.}  
For GemBench, we maintain the learning rate of $8\times10^{-5}$ and AdamW while adjusting the batch size to 160 to accommodate longer trajectories and higher per-sample memory cost. Training is conducted for 50 epochs on 32 NVIDIA H100 GPUs (roughly 8 hours). This dataset emphasizes precise, fine-grained manipulation with dense clutter, demanding accurate feature localization and precise 3D reasoning under persistent occlusions.

\noindent\textbf{Real-robot Fine-tuning.}  
In the final stage, we adapt the model to real-world data using a smaller learning rate of $2\times10^{-5}$ with AdamW and a batch size of 192. Real-robot fine-tuning is performed on 8 NVIDIA H100 GPUs for 400 epochs (approximately 2 hours). This step calibrates ActiveVLA’s viewpoint selection and 3D zoom-in modules to real sensor noise, illumination variations, and complex, non-procedural occlusion patterns encountered during physical execution.

\noindent\textbf{Training Stability and Optimization Notes.}  
Across all stages, we apply gradient clipping at a maximum norm of 1.0 and use cosine learning rate decay. Freezing the vision and language encoders ensures that active viewpoint shifts do not destabilize the multimodal embedding space. Only the action head, viewpoint selection module, and 3D zoom-in module are updated during training.

\begin{figure*}[t]
	\centering
	\includegraphics[width=\linewidth]{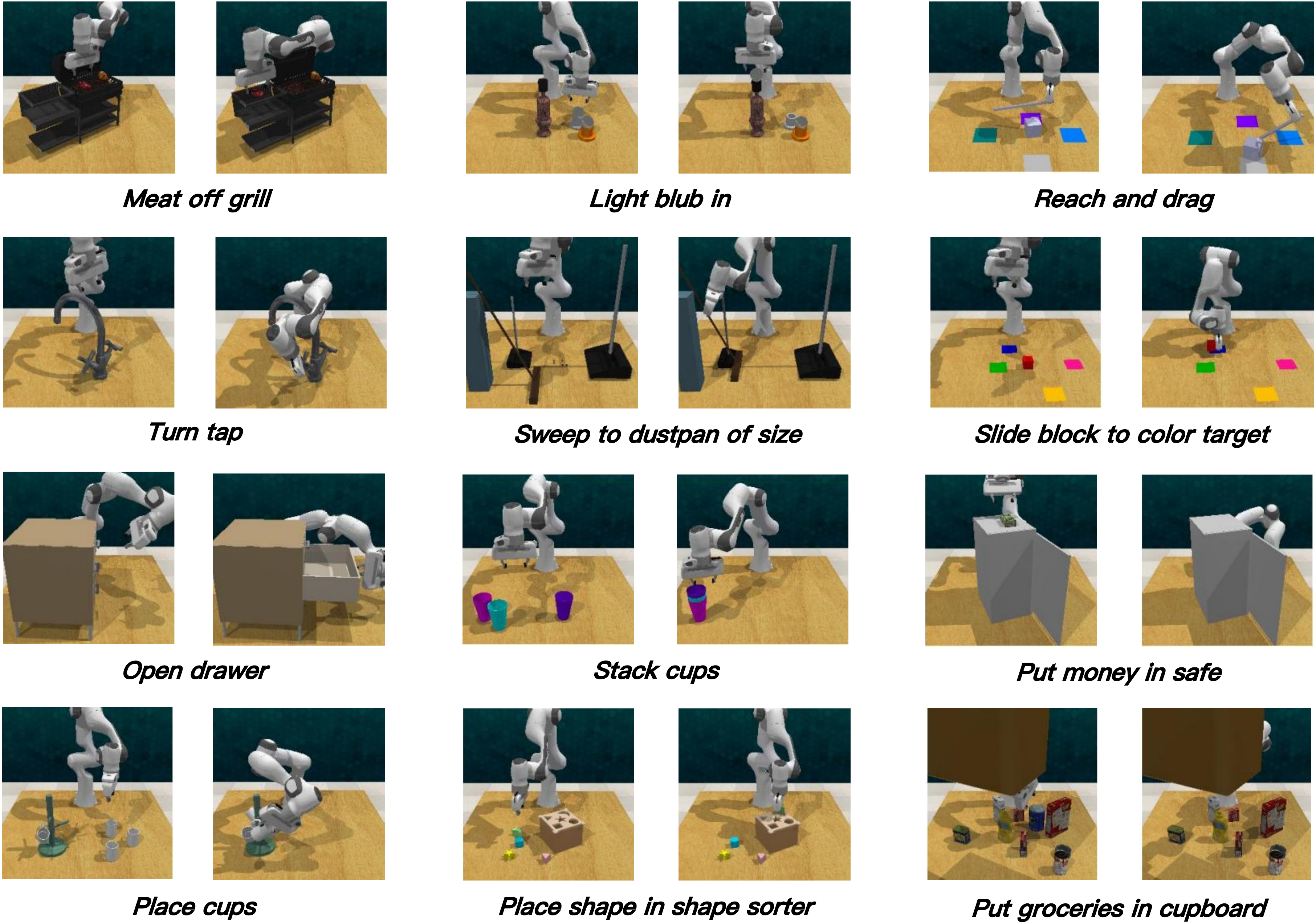}
    \vspace{-0.2in}
    \caption{\textbf{Visualization of 18 RLBench tasks.} 
    The tasks cover a diverse set of manipulation challenges, including object grasping, tool use, and complex spatial interactions, illustrating the range of scenarios used for evaluation. \label{supp_qual_1}}
\end{figure*}

\section{Details of Simulation Tasks}
For the simulation, ActiveVLA is evaluated across three major benchmarks, covering a broad distribution of manipulation difficulty, task compositionality, and robustness challenges. As shown in Figure~\ref{supp_qual_1}, these benchmarks include structured multi-step manipulation, perturbation-heavy robustness evaluation, and hierarchical skill composition, collectively forming a comprehensive simulation suite. This section provides a detailed overview of these simulation domains. The experimental results are shown in Table~\ref{tab:results_activevla}.

\noindent\textbf{RLBench.}  
RLBench serves as the foundational simulation benchmark for ActiveVLA, consisting of 18 visually and physically diverse manipulation tasks executed using a 7-DoF Franka Panda manipulator. Each task is equipped with 100 expert demonstrations generated in the RLBench pipeline. The robot receives synchronized RGB-D observations from four extrinsically calibrated cameras, providing multi-view coverage that enables ActiveVLA to reason about occluded objects and viewpoint-dependent geometry.  
RLBench tasks span fine-grained pick-and-place operations, articulated-object interactions, and multi-stage routines requiring precise control of contact transitions. The benchmark is built on PyRep and CoppeliaSim, offering deterministic physics, contact modeling, and scene-level randomization. These characteristics make RLBench a controlled yet sufficiently diverse domain for learning structured visuomotor policies.

\noindent\textbf{COLOSSEUM.}  
COLOSSEUM extends RLBench into a perturbation-rich robustness benchmark by introducing 12 systematic variation types across object geometry, material, scene layout, kinematics, lighting, and camera configuration. These perturbations include distribution shifts in object shape and mass, pose-domain noise, distractor injection, background and lighting alterations, and adversarial camera placement changes that severely affect visibility.  
Each perturbation type is procedurally generated with controlled parameter ranges, enabling a quantitative evaluation of robustness under viewpoint changes and occlusion-induced ambiguity. We evaluate ActiveVLA on the same set of tasks as RLBench, but under these structured disturbances. Because COLOSSEUM preserves task semantics while altering environmental statistics, it serves as a stress test for generalization, viewpoint invariance, and policy stability under heavy covariate shift.

\begin{figure}[t]
    \centering
    \includegraphics[width=\linewidth]{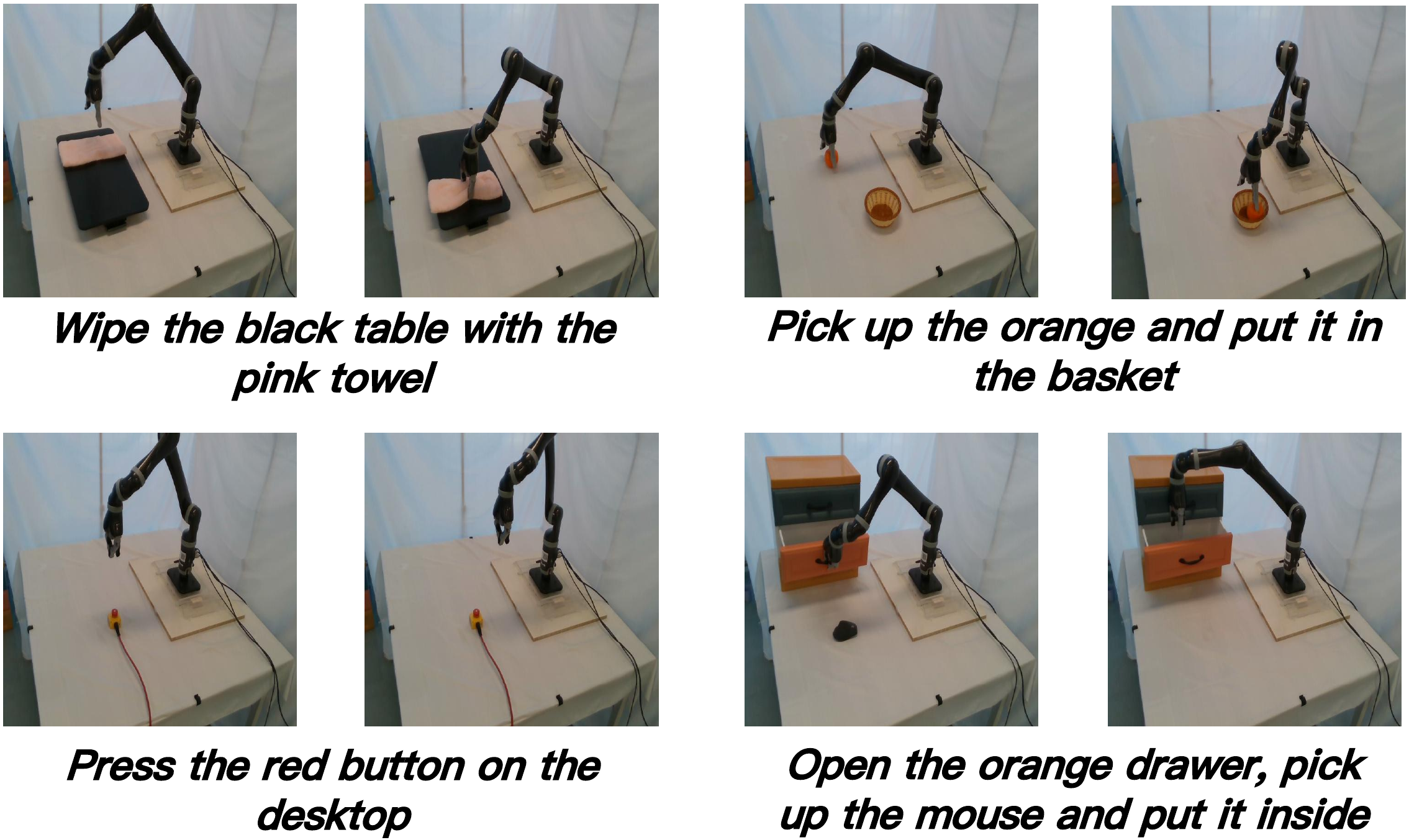}
    \vspace{-0.2in}
    \caption{\textbf{Visualization of direct manipulation tasks.} These tasks evaluate baseline visuomotor skills and include cloth manipulation, highlighting varying physical and spatial challenges.\label{supp_qual_real_1}}
\end{figure}

\begin{figure*}[t]
    \centering
    \includegraphics[width=\linewidth]{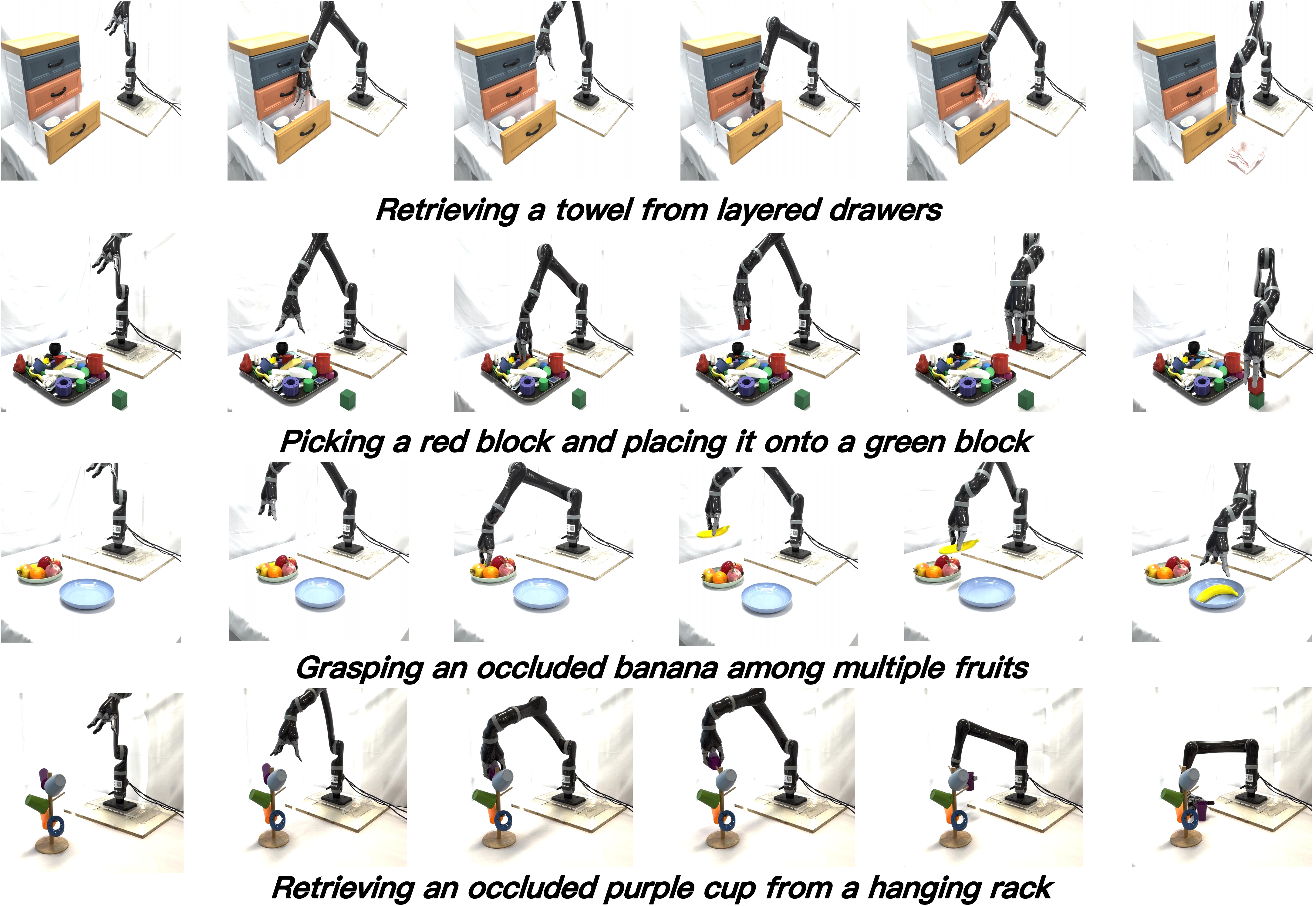}
    \vspace{-0.2in}
    \caption{\textbf{Visualization of occlusion-heavy and spatially complex tasks.} These tasks challenge ActiveVLA’s active viewpoint selection and 3D zoom-in capabilities. They include multi-layered drawer retrieval (\textit{towel}), high-precision stacking with partial occlusion (\textit{red to green block}), collision-aware grasping in clutter (\textit{occluded banana}), and narrow-grasp extraction from a hanging rack (\textit{purple cup}), highlighting severe occlusions, complex geometry, and fine-grained 6D pose reasoning.\label{supp_real_2}}
\end{figure*}

\noindent\textbf{GemBench.}  
GemBench is a hierarchical generalization benchmark built upon RLBench. It decomposes manipulation into seven core action primitives (e.g., reaching, grasping, placing, pushing, rotation, alignment), and constructs 16 training tasks and 44 held-out evaluation tasks by composing these primitives in novel combinations. Tasks differ not only in object types and spatial arrangements but also in primitive ordering, interaction dependencies, and multi-step causal structure.  
Unlike RLBench and COLOSSEUM, where each task follows a predefined demonstration distribution, GemBench requires learning from primitive-level structure and recombining them to solve previously unseen compositions. This setting challenges the model’s ability to perform long-horizon reasoning, adapt to unfamiliar task graphs, and maintain consistent visuomotor grounding across compositional variations.  
The benchmark uses the Franka Panda robot under RLBench physics, with additional environment-induced occlusions and clutter not present in the base tasks. Evaluating on GemBench allows us to measure ActiveVLA’s capacity for skill recomposition, cross-task transfer, and manipulation generalization at the structural level.

\section{Details of Real-Robot Tasks}
We evaluate ActiveVLA on a diverse set of real-robot manipulation tasks that span basic object handling, articulated-object interaction, non-prehensile control, and complex occlusion-heavy scenarios. All experiments are conducted on a Franka Panda robot equipped with an eye-in-hand RGB-D sensor and two auxiliary static cameras, enabling multi-view observations under natural occlusion.

\noindent\textbf{Direct Manipulation Tasks.}  
As shown in Figure~\ref{supp_qual_real_1}, the first group consists of relatively straightforward manipulation tasks designed to assess baseline visuomotor competence:
\begin{itemize}
    \item \textit{Wiping the table with a towel}: The robot grasps a deformable pink towel and wipes a specified region of a black tabletop. Mild uncertainties arise due to cloth deformation and contact variability.
    \item \textit{Picking up an orange and placing it into a basket}: The robot must reliably grasp a spherical object and place it into a confined container, requiring stable grasp synthesis and precise placement.
    \item \textit{Opening an orange drawer and placing a mouse inside}: This sequential task requires articulated-object manipulation, followed by object retrieval and placement within a restricted drawer volume.
    \item \textit{Pushing a blue bowl in front of a green plate}: A non-prehensile manipulation task that relies on continuous contact and collision-free motion in a cluttered planar environment.
    \item \textit{Pressing a red button on the desk}: A reach-and-interact task requiring accurate end-effector pose control but minimal physical complexity.
\end{itemize}

\noindent\textbf{Occlusion-Heavy and Spatially Complex Tasks.}  
As shown in Figure~\ref{supp_real_2},
The second group contains tasks explicitly designed to challenge ActiveVLA’s active viewpoint selection and 3D zoom-in mechanisms. These tasks include severe occlusion, multi-layer geometry, and fine-grained 6D pose reasoning:
\begin{itemize}
    \item \textit{Retrieving a towel from layered drawers}: The robot must identify the correct drawer, reason about partial occlusions, actuate the drawer, and extract the towel from a tightly constrained multi-layer structure.
    \item \textit{Picking a red block and placing it onto a green block}: A high-precision stacking task where the target object is partially occluded or surrounded by distractors, demanding fine-grained localization and vertical alignment.
    \item \textit{Grasping an occluded banana among multiple fruits}: The banana is frequently hidden beneath or between other fruits, requiring collision-aware grasp planning and viewpoint selection in dense clutter.
    \item \textit{Retrieving an occluded purple cup from a hanging rack}: The rack geometry induces persistent occlusions and narrow allowable grasp regions. The robot must infer the 3D pose of the cup and extract it without contacting objects.
\end{itemize}

These occlusion-heavy tasks pose substantially greater challenges due to compounded visibility constraints, irregular geometry, and the need for precise 6D pose inference.

\end{document}